\documentclass[conference]{IEEEtran}
\usepackage{times}
\usepackage[numbers]{natbib}
\usepackage{multicol}
\usepackage{amssymb}
\usepackage{amsmath}
\usepackage[bookmarks=true]{hyperref}
\usepackage{xcolor}
\usepackage{tabularx}
\usepackage{hyperref}       
\usepackage{url}            
\usepackage{booktabs}       
\usepackage{amsfonts}       
\usepackage{nicefrac}       
\usepackage{microtype}      
\usepackage{xcolor}         
\usepackage{lipsum}         
\usepackage{amsmath}
\usepackage{graphicx}
\usepackage{caption}
\usepackage{lipsum}
\usepackage{stfloats}
\usepackage{subfigure}

\newcommand{\ms}[2]{\small $\mathop{#1}_{\pm #2}$}
\newcommand{\msb}[2]{\small $\mathop{\textbf{#1}}_{\pm \textbf{#2}}$}
\makeatletter
\def\blfootnote{\xdef\@thefnmark{}\@footnotetext}
\makeatother

\begin{document}


\title{Rotating without Seeing: \\ Towards In-hand Dexterity through Touch}


\author{Zhao-Heng Yin$^{1, *\dagger}$, Binghao Huang$^{2, *}$, Yuzhe Qin$^{2}$, Qifeng Chen$^{1}$, Xiaolong Wang$^{2}$ \\
$^{1}$HKUST \hspace{0.2in}
$^{2}$UC San Diego\\
{\color{blue}{\texttt{\url{http://touchdexterity.github.io}}}}
}

\twocolumn[{%
\renewcommand\twocolumn[1][]{#1}%

\maketitle
\begin{center}
    \centering 
    \includegraphics[width=\linewidth]{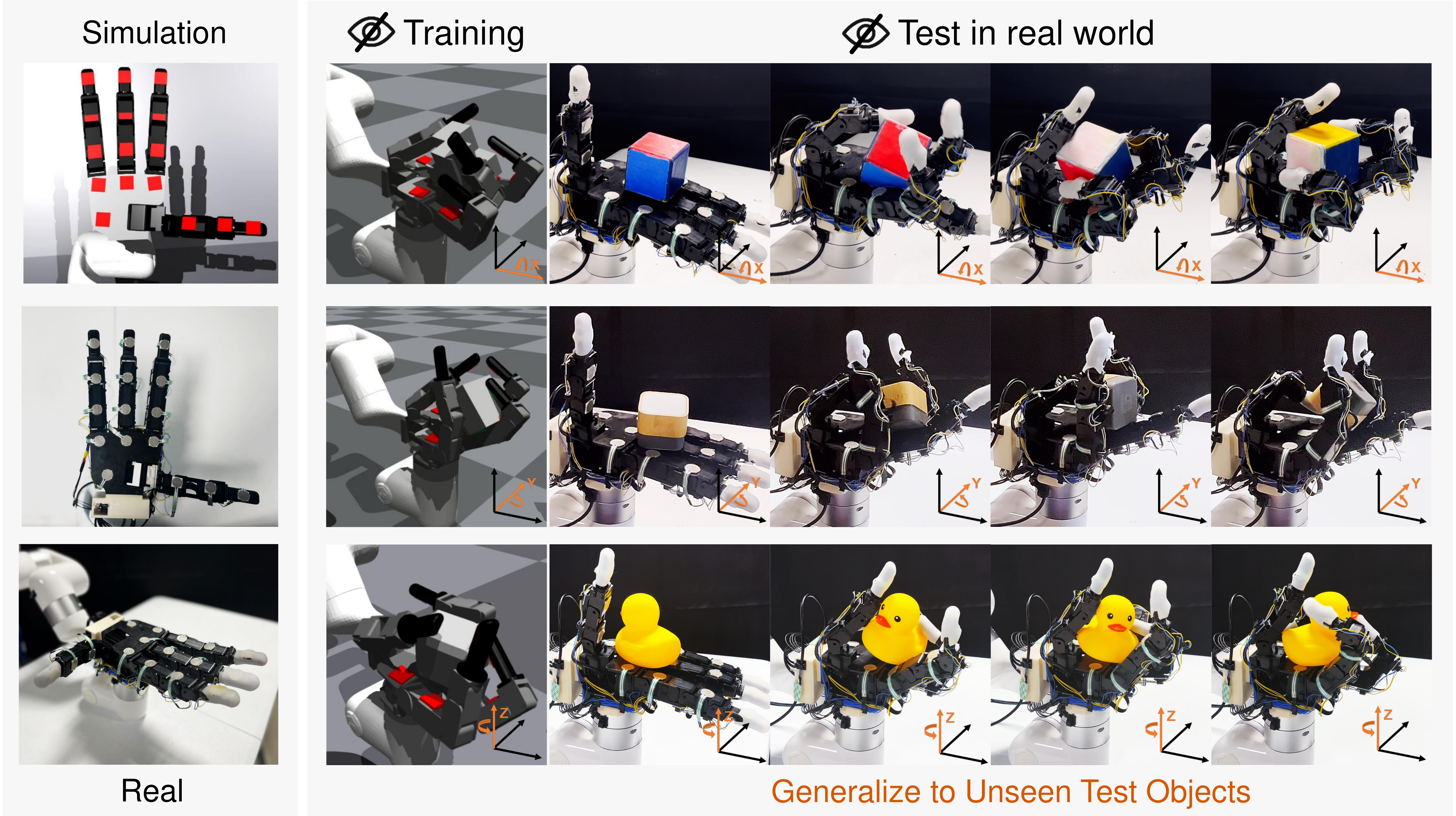}
    \captionof{figure}{We propose \textbf{Touch Dexterity}, a new dexterous manipulation system to perform in-hand object rotation with only touch sensing. On the left, we show our hardware setup with 16 FSR sensors attached to an Allegro hand. We train our policy in simulation on rotating diverse objects around different axes. Our trained policy can be directly transferred to the real robot hand and can rotate novel/unseen objects successfully.}
    \label{fig:teaser}
   
\end{center}
  }]
{\blfootnote{{$^{*}$ The first two authors contributed equally.}}}
{\blfootnote{{$^{\dagger}$ Work done while an intern at UC San Diego.}}}
\begin{abstract}
Tactile information plays a critical role in human dexterity. It reveals useful contact information that may not be inferred directly from vision. In fact, humans can even perform in-hand dexterous manipulation without using vision. Can we enable the same ability for the multi-finger robot hand? In this paper, we present Touch Dexterity, a new system that can perform in-hand object rotation using only touching without seeing the object. Instead of relying on precise tactile sensing in a small region, we introduce a new system design using dense binary force sensors (touch or no touch) overlaying one side of the whole robot hand (palm, finger links, fingertips). Such a design is low-cost, giving a larger coverage of the object, and minimizing the Sim2Real gap at the same time. We train an in-hand rotation policy using Reinforcement Learning on diverse objects in simulation. Relying on touch-only sensing, we can directly deploy the policy in a real robot hand and rotate novel objects that are not presented in training. Extensive ablations are performed on how tactile information help in-hand manipulation.

\end{abstract}

\IEEEpeerreviewmaketitle

\section{Introduction}
Imagine we are washing the used pan in the kitchen after dinner. Suddenly, the power is cut off unexpectedly, and all the lights go out. What would we do? Most of us may stop the work, put down the pan in the sink, and then probably find our phone in the pocket to light up the way. Simple as it may seem, this sequence of actions actually requires precise execution of in-hand dexterous manipulation in the dark, where we receive no vision input for guidance. Even in normal situations with lights on, the manipulation of objects in hand often comes with heavy occlusions. Without relying on vision, we humans are still very good at feeling and manipulating objects by hand, which is made possible by the tactile~(touch) information coming from our skin. Previous studies in biology also confirm the vital importance of touch information for dexterous manipulation~\cite{human_tactile}. Can we enable robots with such dexterity with touch sensing?

Indeed, tactile sensing has been a long-standing topic in robotics. With different designs of tactile sensors, robots are able to manipulate objects more precisely using contact information~\cite{tactile_insertion, kolamuri2021improving_grasp, contact_ar} and even complete tasks in a touch-only setup ~\cite{tactile2020review,murali2020learning}. However, it is still very challenging for touch-only approaches to achieve complex and high degree-of-freedom (DOF) in-hand manipulation. While most current literature focuses on modeling precise and fine-grained contact using increasingly high-quality sensors, it introduces two challenges to in-hand manipulation: (i) Most approaches are only able to attach the expensive sensors to the finger-tips of the gripper or hands instead of covering the whole manipulator, limiting the range of tasks to perform; (ii) It often requires a large number of training samples for complex tasks, but it is hard to leverage a simulator given the Sim2Real gap is usually very large for a delicate sensor. 

In this paper, we present \textbf{Touch Dexterity}, a new system design and learning pipeline for in-hand rotation using only touching. Instead of using a few sensors on finger-tips that give high-quality patterns~\cite{yuan2017gelsight,lambeta2020digit,padmanabha2020omnitact}, we propose the alternative: Use a lot of low-cost binary force sensors (touch or no touch) attached over one side of the hand (fingertips, links, and palm) as shown in Fig.~\ref{fig:teaser} (left). Specifically, we attach the Force-Sensing Resistor (FSR) sensors, which cost around $\$12$ each on Amazon\footnote{\url{https://www.amazon.com/s?k=fsr+sensor}}, to the Allegro robot hand. Our insight is that, while one single binary force sensor cannot do much, the combination of 16 of them has a strong representation power ($2^{16}$ types of states in maximum), which might allow the robot hand to ``feel'' the object state without seeing. Importantly, the Sim2Real gap by using such a binary sensor is minimized to the extreme, which allows large-scale sample collection in simulation for training. 

With this system setup, we focus on the task of rotating an ``unseen'' object around the $x$, $y$, and $z$-axis using the multi-finger hand as shown in Fig.~\ref{fig:teaser} (right). Here ``unseen'' not only indicates there is no vision, but also means the object is not presented during training time. While this task is a simplified version of the in-hand re-orientation task, it is still very challenging as all the fingers are moving with a relatively large motion to rotate the object and prevent it from falling off the palm at the same time. We believe the same pipeline can be directly extended to more complex tasks in the future. We train our policy on multiple objects in parallel in the IsaacGym simulator~\cite{isaacgym} using Reinforcement Learning~(RL), and the learned policy can be directly deployed on the real robot manipulating diverse unseen objects. The key to achieving such generalization across objects and to the real robot is our touch sensors. Our RL policy takes both the binary touch sensing information and the robot's internal state as input and predicts the action in each time step for closed-loop control. With a large coverage over the object using the touch sensors, our hypothesis is that the policy implicitly learns to understand the 3D structure and pose of the object and perform rotation accordingly. 

In our experiments, we test the real-world system with 10 diverse objects. Our method shows surprising robustness in rotating unseen objects using only touch sensing. For example, we can rotate the rubber duck for two cycles without falling, even if it is never presented in training (last row in Fig.~\ref{fig:teaser}). We perform extensive ablations on our sensor to validate our design, including disabling all the touch sensors, disabling part of them, and using continuous signals instead of binary signals. 

\section{Related Work} \label{relatedworks}

\noindent \textbf{Dexterous Manipulation} Dexterous manipulation has been a long-standing problem in robotics~\cite{rus1999hand, dex_overview, sliding_dex, rolling_dex, pivot_dex, dex_1, dex_2, dex_3, dex_4, dex_5, dex_6, dex_7, dex_8, rl-openai}. Among these works, dexterous in-hand manipulation receives a lot of attention in recent years~\cite{dexhand_1_kumar2014real_time_behaviour_synthesis, dexhand_2_Bhatt2021surprisingly, dexhand_3_bai2014dexterous, dexhand_4_morgan2022complex, rl-openai}. Several early methods propose to tackle the in-hand manipulation problem with analytical model-based approaches~\cite{dexhand_1_kumar2014real_time_behaviour_synthesis, dexhand_3_bai2014dexterous}. Nevertheless, they pose certain hypotheses about the objects and the controllers, which makes it hard to scale to more complex tasks. To overcome this limitation, deep Reinforcement Learning has been applied recently on dexterous manipulation~\cite{rl-openai, dextreme, corl21-inhand, qin2022dexpoint, chen2022visual, qi2022hand}. Building on these works, incorporating demonstrations in with imitation learning also leads to better sample efficiency and more natural manipulation behaviors~\cite{qin2022dexmv, dapg, arunachalam2022dexterous, qin2022one, wu2023learning, ye2022learning, liu2022herd, patel2022learning, arunachalam2022holo}. However, most in-hand manipulation methods are still highly relying on visual inputs~\cite{rl-openai,dextreme,chen2022visual}. For example, Chen et al.~\cite{chen2022visual} propose to perform in-hand object re-orientation  using depth image input, and new hardware is designed to avoid heavy occlusion. Instead of relying on vision which faces the occlusion problem with general hardware, recently, several works~\cite{qi2022hand, Sievers2022, Pitz2023} propose to perform in-hand object rotation without both visual and explicit tactile sensing. The idea of these works is that we can infer the object's information from the implicit tactile information inside proprioception data. However, these works either only consider object rotation on the fingertip with relatively small finger motion, or the rotation of a limited set of objects. Compared to these works, our system explicitly use touch sensors to percept hand-object interaction, and can solve the object rotation problem on the palm for diverse types of objects, which involves complex object motion and is more challenging. We also find that using explicit tactile sensing enables our touch-based policy to generalize to unseen objects, which is not shown by previous works. 
 
\noindent \textbf{Tactile Robotic Manipulation} Biological evidence suggests that tactile information is crucial for the success of human dexterity~\cite{human_tactile}. This basic observation naturally motivates the research of tactile robotic manipulation~\cite{6943031, molchanov2016contact, tactile_insertion, bhattacharjee2015material, navarro2015active, zhu2022learning, roller_tactile2022, suresh2022midastouch, tactile_slam, tandem, object_track2022patchgraph, visiotactile, manipulation_by_feel, smith2021active, tactile2020alberto, roller_tactile2022, murali2020learning, taylor2022gelslim,https://doi.org/10.48550/arxiv.1907.13098,gao2022objectfolder,bhirangi2021reskin}. A fundamental question is what kind of touch information is essential. Existing works propose to extract local geometry, force and torque, contact event, and material properties with various sensors to help manipulation~\cite{tactile2020review}. Different from these works, we find that even using the simplest binary contact signal provided by a sparse sensor array can be helpful for a high-dimensional manipulator. This is also found in~\cite{bin_contact_1, bin_contact_2, liang2020hand} where binary contact signals are used for manipulation, object tracking and exploration. However, they focus on low-DOF manipulators rather than a multi-finger robot hand. In the dexterous hand research, Buescher et al.~\cite{tactile_skin} develop a skin-based tactile sensing system on the Shadow hand, which has a similar but denser sensor layout over the palm compared with our work. However, it is still unclear how to use it with a control method to solve  in-hand rotation as in our work. Another important question in tactile robotic manipulation is how to simulate the tactile event so as to perform Sim2Real transfer. Researchers have proposed many approaches and strategies for tactile simulation~\cite{moisio2013model,habib2014skinsim,tactile_sim,church2022tactile,si2022taxim,bi2021zero}. For example, Xu et al.~\cite{tactile_sim} propose a method to simulate normal and shear tactile force field on the contact surface. Compared with these works, our method does not require any extra simulation design but can leverage the built-in contact simulation of an existing physics simulator. 

\section{Tactile Dexterous Manipulation System}

\subsection{Real-word System Setup}
Our hardware setup consists of a XArm robot arm and a 16-DOF Allegro Hand with a contact sensor array. The array consists of 16 contact sensors, which are attached to different parts of the allegro hand including the palm and tips as shown in Figure~\ref{fig:teaser}(left). The used contact sensors are based on Force-Sensing Resistors (FSR), whose resistance will change when an external force is applied to its surface. These sensors are very sensitive to force and widely used in robotics. We use an STM32F microcontroller to collect the analog voltage signals of each sensor and then forward digital signals to the host.

While these contact sensors are able to output the \textbf{continuous} contact force measurement, the signals are usually nonlinear and noisy. As a result, it should undergo necessary preprocessing before being used for control. We \textbf{binarize} these measurements with respect to a selected threshold $\theta_{th}$ and use this binary contact signal for control. The advantage of using binary signals is that it can reduce the gap between the simulation and the real robot, and simplify the Sim2Real transfer procedure. When using the exact force measurement as observation, it is difficult to align the measurement between the simulation and the real robot, especially since there are still errors in aligning the analog voltage signals to the exact force measurement. In contrast, we can easily calibrate the binarized measurement by adjusting the threshold.

\subsection{Simulation Setup}

In this paper, we use the IsaacGym simulator~\cite{isaacgym} for the training of our tactile manipulation system. The simulation setup is shown in Figure~\ref{fig:teaser}(left). We simulate each contact sensor as a fixed link on the finger and palm links. We fetch the net contact force $F=[F_x, F_y, F_z]$ over each sensor link provided by the simulator at each simulation step, and use $\Vert F\Vert$ as the simulated contact force measurement. Then, we binarize the measurement with another threshold $\tilde\theta_{th}$, Note that the force provided by the sensor's parent link does not contribute to the net contact force. We adjust the threshold $\tilde\theta_{th}$ of these sensors to ensure that they have similar behavior to that in the real. We use a $\tilde\theta_{th} = 0.01N$ in simulation.
\begin{figure}[t]
    \centering 
    \includegraphics[width=\linewidth]{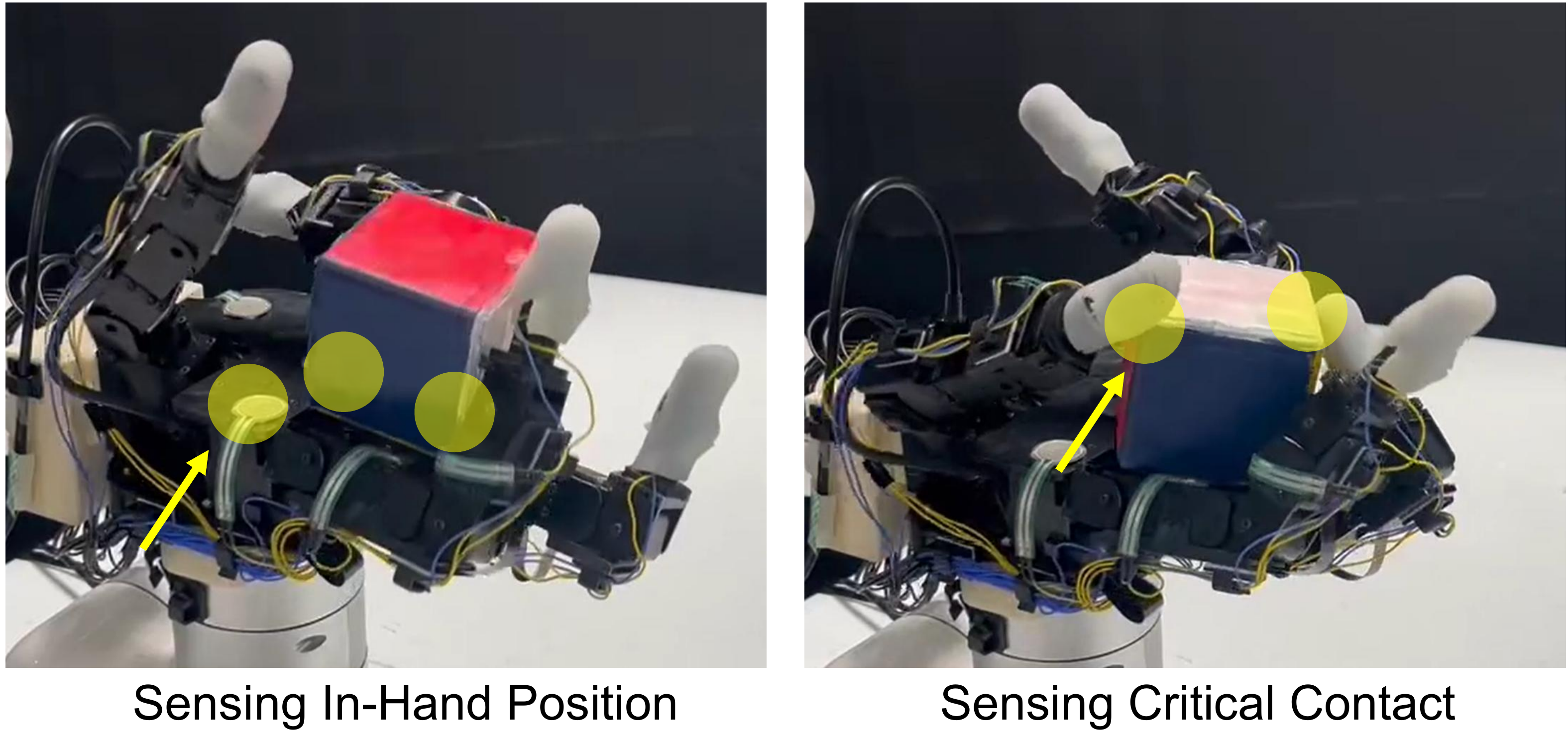}

    \caption{Two major functionalities of our sensors: sensing (i) the objects' in-hand position, and (ii) the critical contact during the dexterous manipulation process. Note that we use finger cots to increase the friction and we still have force-sensing resistors inside the finger cots.}
    
    \label{fig:setup_and_function}
    \vspace{-0.1in}
\end{figure}
\subsection{Benchmark Problem: In-hand Rotation}
In this paper, we study the dexterity of our system by using it to solve an in-hand rotation task. In this in-hand rotation task, an object is initialized in the palm and the robot hand is then required to rotate this object around a given rotation axis. 

When we are doing in-hand object rotation, the object motion is more complex than that in finger-tip rotation mentioned in section ~\ref{relatedworks} and brings additional challenges. Specifically, the object can slide or roll in the palm during in-hand manipulation. Due to this complex motion pattern, explicit feedback from tactile or vision becomes necessary for successful manipulation. Otherwise, we are unable to infer the current state of the object and fail to push and rotate it in a secure way.

\subsection{Discussion: What information can sensors provide?}
We summarize two kinds of information our system can provide for control as follows, though its sensing is sparser than that of a real human hand.

\textbf{Position information.} The contact sensors can inform the policy where the object is at each time step. One example is shown in Figure~\ref{fig:setup_and_function} (left). In this example, a cuboid is placed on the palm without contacting any fingertip. At this moment, the only way to infer the position of the object is by reading the measurement of the contact sensors on the palm. This measurement can provide an estimation of the object's position~(i.e., at the center), based on which the controller can decide the approximate movement of each finger, for example, driving the thumb toward the center. Without this information, the thumb can move to the right and can not come into contact with the object to initiate a rotation. 

\textbf{Interaction information.} During in-hand object rotation, it is essential to ensure that the fingertip in charge of the rotation is indeed interacting with the object, see Figure~\ref{fig:setup_and_function} (right). Otherwise, the finger may not be able to push against the object leading to a failure, which may cause the object to move to an unstable position, and even fall out of the hand.

\begin{figure}[t]
    \centering 
    \includegraphics[width=\linewidth]{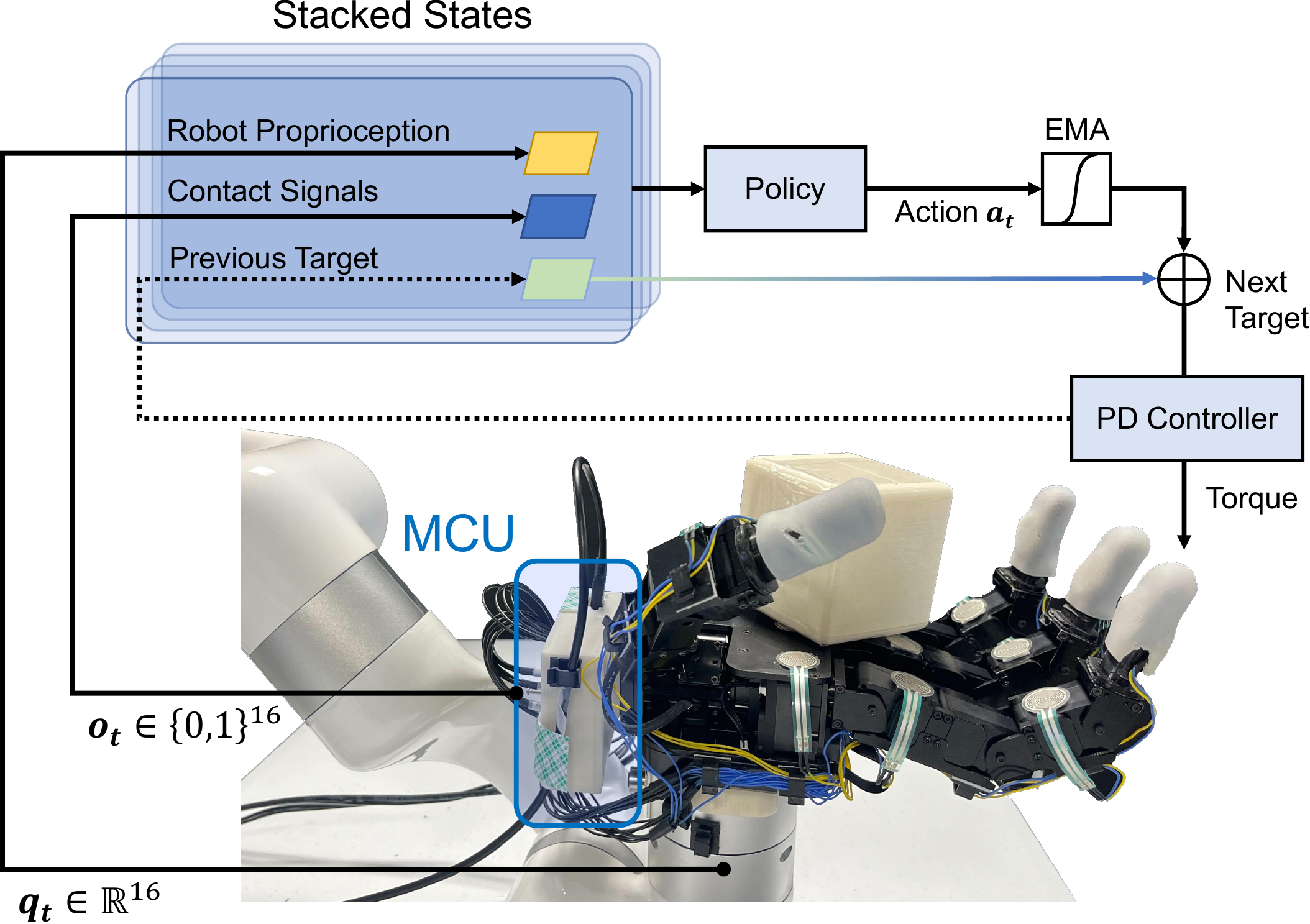}
    \caption{Overview of the control process. The state contains tactile information, joint position, previous target, and task information like rotation axis~(not shown in the figure). The policy then uses the stacked state to get the relative action, and the next target joint position is calculated. The new target is then fed to a PD controller.}
    \label{fig:realpipeline}
    \vspace{-0.1in}
\end{figure}

\section{Learning Touch Dexterity}

\subsection{Problem Formulation}
We formulate the in-hand rotation problem as a Markov Decision Process $\mathcal{M}=(\mathcal{S}, \mathcal{A}, \mathcal{R}, \mathcal{P})$. Here, $\mathcal{S}$ is the state space, $\mathcal{A}$ is the action space, $\mathcal{R}$ is the reward function, and $\mathcal{P}$ is the transition dynamics. $\mathcal{R}$ and $\mathcal{P}$ are unknown to the robot. The robot agent observes state $s_t$ at each step $t$ and take action $a_t = \pi(s_t)$ calculated by the current policy $\pi$, then it will receive a reward $r_t = \mathcal{R}(s_t, a_t, s_{t+1})$. The goal of the agent is to maximize the $\gamma$ discounted return $\sum_{t=0}^T\gamma^t r_t$. The definition of these elements is as follows.

\subsubsection{State}
The state of the system consists of the joint position of the Allegro hand $q_t\in\mathbb{R}^{16}$, the sensor observation $o_t \in \{0, 1\}^{16}$, the previous position target $\tilde{q}_{t}\in\mathbb{R}^{16}$, and the rotation axis $k\in\mathbb{S}^2$. Since the state at one step may not be sufficient for control, we also stack it with other 3 historical states as the input when we use an MLP as the policy network.

\subsubsection{Action} At each step, the action produced by the policy network is a relative control command $a_t\in\mathbb{R}^{16}$. A PD controller then drives the hand to reach the joint position target $\tilde{q}_{t+1} = \tilde{q}_{t} + a_t$ at the next step. However, using this target directly may lead to non-smooth finger motion, since the actions of two consecutive steps may conflict with each other. Therefore, in practice, we use an exponential moving average as the target: $\tilde{q}_{t+1} = \tilde{q}_{t} + \tilde{a}_t$, where $\tilde{a}_t = \eta a_t + (1 - \eta)\tilde{a}_{t-1}, t\geq 1$ and $\tilde{a}_0 = 0$. We find that $\eta = 0.8$ works well in the experiments. This PD controller operates at a control frequency of 10Hz both in the simulation and the real.

\subsubsection{Reward} 
\begin{figure}[t]
    \centering 
    \includegraphics[width=0.85\linewidth]{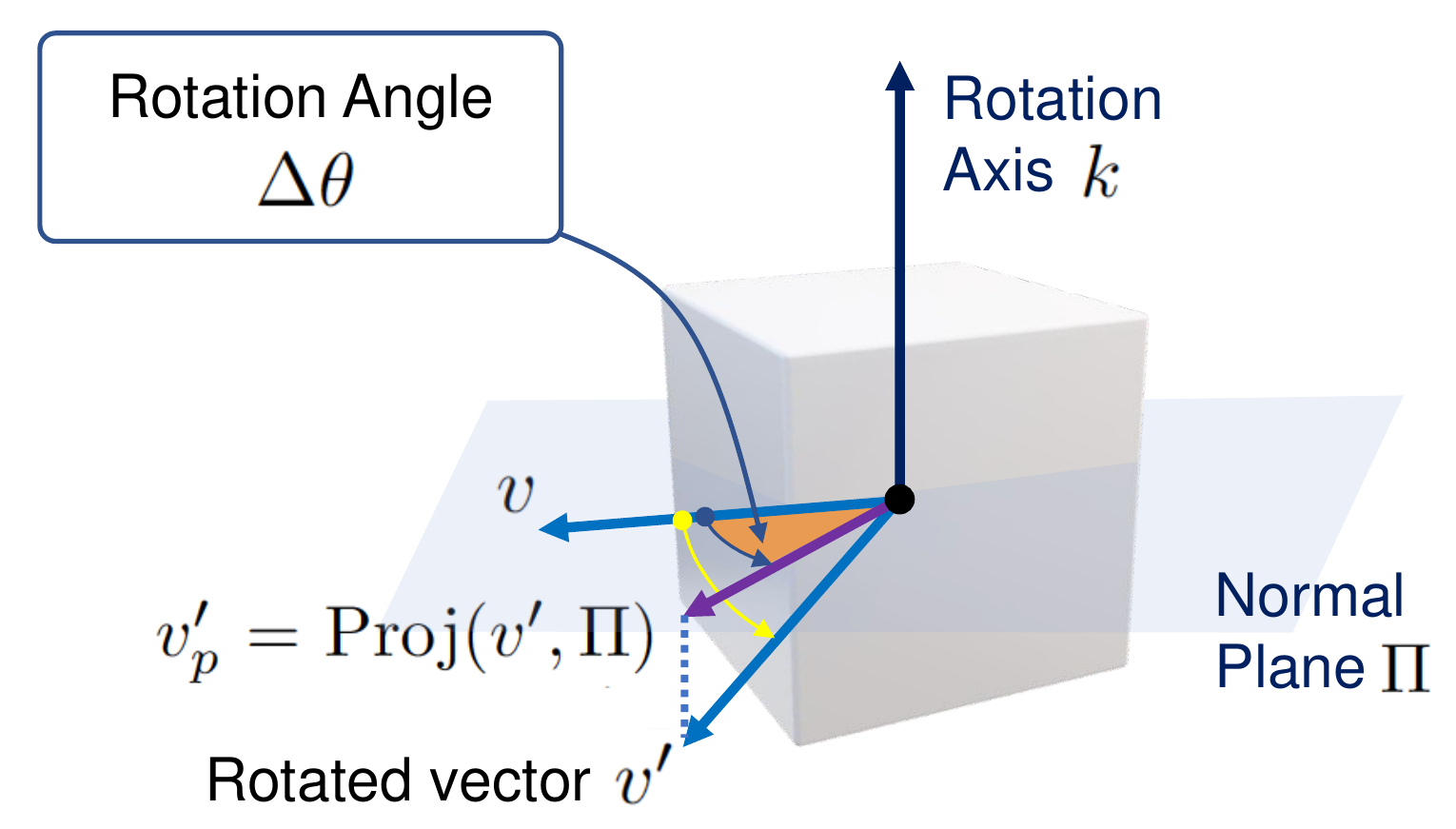}
    \caption{Illustration of the calculation of rotation angle $\Delta \theta$: The object rotates alone Axis $k$ and here we visualize the rotation angle $\Delta \theta$ in the Normal Plane.}
    \label{fig:rot}
    \vspace{-0.1in}
\end{figure}

We design a reward function that is able to make the dexterous hand rotate the object in a smooth and transferable way. The reward function used in this paper is a weighted mixture of several components:
\begin{equation}
    r_t = w_1r_{rot} + w_2r_{vel} + w_3r_{fall} + w_4r_{work} + w_5r_{torque} + w_6r_{dist}.
\end{equation}
The first term $r_{rot}$ is the rotation reward defined as the rotated angle $\Delta \theta$ of a sampled unit vector in the normal plane $\Pi$ of the rotation axis $k$:
\begin{equation}
    r_{rot} = {\rm clip} (\Delta \theta, -c_1, c_1).    
\end{equation}
The detailed calculation of $\Delta \theta$ is shown in Figure~\ref{fig:rot}. First, we sample a unit vector $v$ in $\Pi$ randomly and we may as well imagine it is attached to the object. Then we fetch its corresponding vector $v'$ at the next state and project it to $\Pi$: $v'_p =  {\rm Proj} (v', \Pi)$. $\Delta \theta\in [-\pi, \pi)$ is defined as the signed distance between $v'_p$ and $v$ with respect to the axis $k$. Note that \cite{qi2022hand} uses $\langle \omega, k\rangle$ as the rotation reward, where $\omega$ is the angular velocity returned by the simulator. Nevertheless, we find that the angular velocity provided by the simulator in our setting is very noisy since the motion of the object is very complex. As a result, using this angular velocity in the reward can usually lead to very undesirable object motion patterns, like vibrating around a specific pose. We find that using this finite difference as the reward can produce consistent rotation behavior across different runs. The second term is a penalty on the object's velocity $r_{vel} = -\Vert v_t\Vert$. This encourages the hand to rotate the object in a stable manner and increases the transferability of the trained policy. The third reward $r_{fall}$ is a negative falling penalty when the object falls out of the palm. The fourth reward $r_{work}$ penalize the work of controller, which is defined as $r_{work} = -\langle|\tau|, |\dot{q_t}|\rangle$. Here, $\tau$ is the outputted torque of the PD controller at step $t$. This penalty helps to improve the smoothness of finger motion. The fifth term $r_{torque} = -\Vert\tau\Vert$ penalizes the large torque. Finally, $r_{dist} = {\rm mean}({\rm clip} (1/(\epsilon + d(x_{tip}, x_{obj})), c_2, c_3))$ is a distance reward, which encourages the fingertip to come close to the object and interact with it.

\subsubsection{Reset Strategy} We design several reset strategies to reduce unnecessary exploration and speed up the learning process. First, we reset the episode when the object deviates too much from its initial position~(i.e., the center of the palm). Moreover, we reset the episode when the major axis of the object deviates too much from the rotation axis, this reduces the exploration of an undesired rotation direction.

\subsection{Domain Randomization}
We use a wide variety of domain randomization~\cite{domain_randomization} to improve the Sim2real transfer.

\subsubsection{Physics randomization} We randomize the object's initial position, mass, shape, and friction to ensure that the learned policy can deal with different kinds of objects. 

Moreover, we randomize the gain of the PD controller to model the uncertainty of the PD controller in real. Besides, we consider randomizing each tactile sensor. For each activated contact sensor that outputs 1, with probability $p$ we flip its output to 0. We also model the signal delay of the contact sensor by an exponential delay used in~\cite{dextreme}.

\subsubsection{Non-physics randomization} 
We use a set of non-physics randomization to further improve the robustness of the trained policy. We inject white noises into the observation of the policy, and its outputted action to ensure that it is robust to small perturbations.

\subsection{Training Procedure}
We use the proximal policy optimization~(PPO)~\cite{ppo} algorithm to train our control policy and multilayer perceptron~(MLP) for both of the policy and value networks. We use the advantage clip threshold $\epsilon=0.2$ and the KL threshold of 0.02. We use ELU~\cite{relu} as the activation function in these networks. The policy network outputs a Gaussian distribution with a learnable state-independent standard deviation. Like~\cite{dextreme}, in order to reduce the training difficulty, we also use asymmetric observation for the policy and value network. Concretely, for the value network, we add privileged information such as the contact force over each link, the object's ground-truth pose, and physical parameters to its input. This privileged information is not accessible by the policy network. For the policy network, we only stack the current state with 3 historical states as the input. 

For the IsaacGym simulation, we set $dt=0.01667s$ with 2 simulation substeps. We use 8192 parallel environments. The action~(control target) outputted by the policy network is executed by 6 steps, corresponding to a 10Hz control frequency in real.
\begin{figure}[t]
    \centering 
    \includegraphics[width=\linewidth]{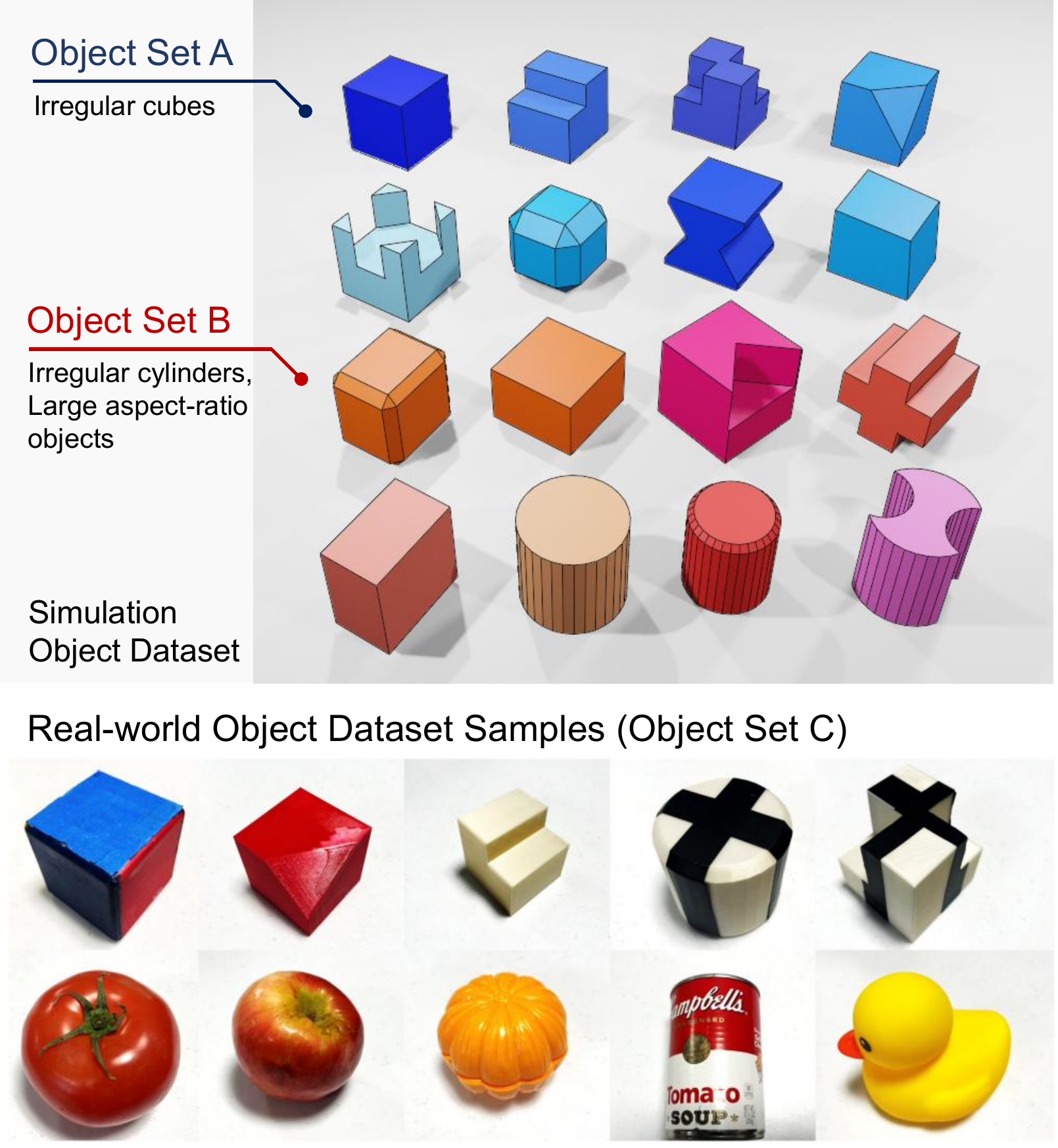}
    \caption{The object sets used in our experiments. The full object set in the real world can be found in the supplementary material.}
    \label{fig:object}
    \vspace{-0.1in}
\end{figure}

\section{Experiments}
In this part, we compare our Touch Dexterity system to several baselines in both the simulation and the real. Specifically, we are interested in the following questions:
\begin{enumerate}
    \item \textit{How much benefit does tactile information offer compared to the baseline in training?} 
    \item \textit{Using simulation as an ideal setup, does the usage of tactile information lead to better robustness and generalization?} 
    \item \textit{How well does our tactile manipulation system perform and generalize compared with other methods in the real?}
    \item \textit{How well does tactile perception in simulation align with that in real? How does it improve performance in real?}
\end{enumerate}
We answer these questions through an extensive case study on the $z$-axis rotation. Then, we demonstrate that our system can also learn the rotation skill along all the other axes.
\subsection{Experiment Setup}
\subsubsection{Object Dataset}
For the simulation experiments, we train and evaluate our policy on a set of artificial objects of common geometries, such as cuboids, cylinders, and balls. Some examples of these objects are shown in Figure~\ref{fig:object}. Despite their simplicity, their diverse geometry can be used to approximate a large set of common daily objects. For the real experiments, we bring in some unseen real-world objects like a rubber duck, lego box for evaluation as shown in Figure~\ref{fig:object}. 
\subsubsection{Evaluation Metric}
To evaluate the performance of a trained policy, we introduce the following metric as suggested by~\cite{qi2022hand}.
\begin{enumerate}
    \item \textbf{Cumulative Rotation Reward~(CRR).} We calculated the cumulative rotation reward to evaluate the rotation capability of a policy in the simulation. This metric is only used in the simulation. 
    \item \textbf{Cumulative Rotation Angle~(CRA).} We count the cumulative rotation angle~(by rounds) to evaluate the rotation capability of a policy in the real. This metric is counted by a human.
    \item \textbf{Time-to-Fall~(TTF/Duration).} We measure the time~(by seconds) of an object staying in the palm before falling down the hand. This metric can be used both in the simulation and in real. 
\end{enumerate}
\begin{figure*}[ht]
\subfigure{\includegraphics[width=\textwidth]{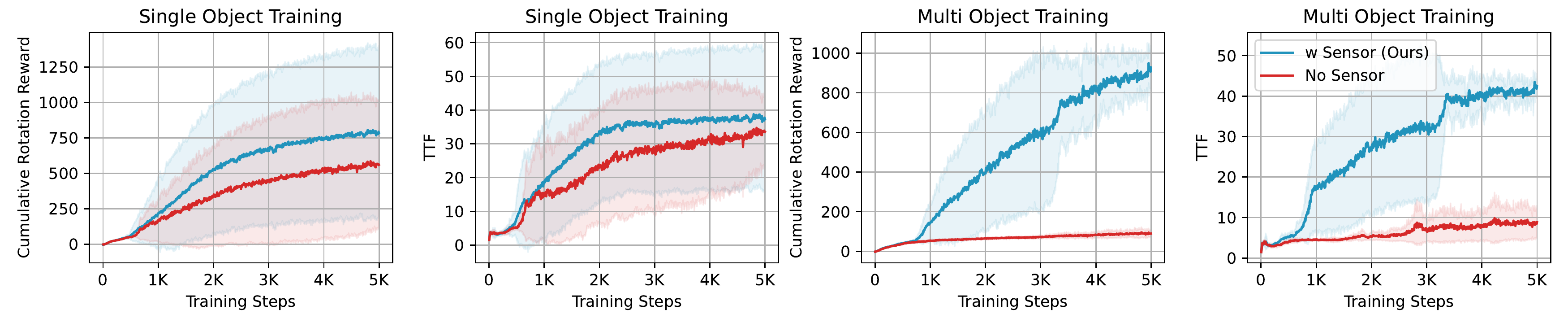}}
\subfigure{\includegraphics[width=\textwidth]{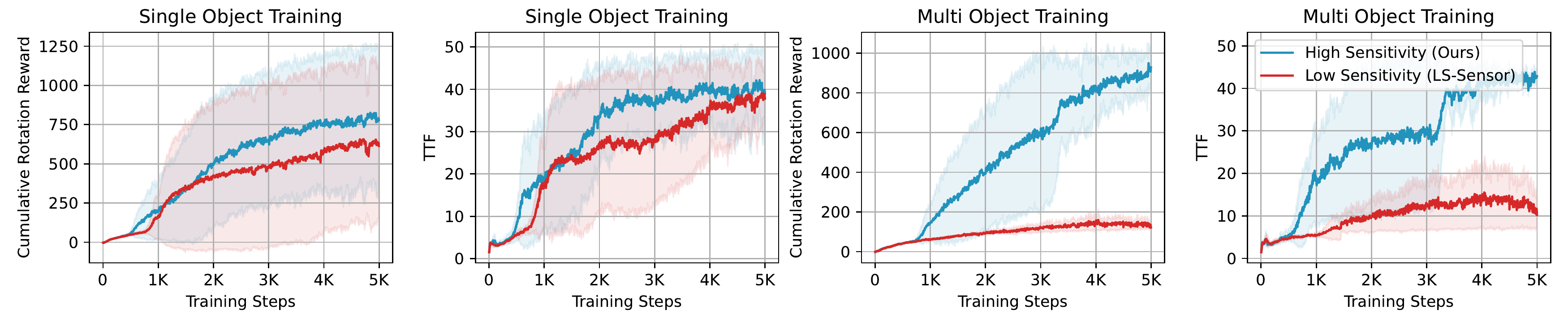}}
\vspace{-0.2in}
\caption{Top: Policy training curve with and without sensors. Bottom: Policy training curve with sensors of different sensitivities. The results are averaged on 3 seeds. The shaded area shows the standard deviation.}
\label{fig:curve}
\vspace{-0.4cm}
\end{figure*}

\subsection{Baselines}
In the experiments, we mainly compare our methods with the following baselines. 
\begin{enumerate}
    \item \textbf{No-Sensor.} We train a PPO policy to control the hand with no tactile information available. The only way to infer the object-hand interaction information is by comparing the current joint position to the desired, target joint position. For example, when a finger~(e.g. thumb) is pressing the top surface of a cuboid, we can observe a difference between these two quantities, indicating the existence of pressing behavior.
    \item \textbf{LS-Sensor.} We set a higher sensor activation threshold $\theta_{th}=0.2N$ and train another PPO policy. In other words, the sensors now have lower sensitivity to the contact, and we call this policy LS-Sensor. Under this setup, the hand is no longer able to sense some slight contacts.
    \item \textbf{DS-Sensor.} This policy is used for ablation purposes. Its only difference from our policy is that it will disable all the tactile input during evaluation. This is used to test to which extent the trained tactile policy uses tactile information.
\end{enumerate}
In the real-world experiments, we also introduce additional two policy baselines: 
\begin{enumerate}
    \item \textbf{Openloop Policy.} We collect several successful object rotation trajectories in the simulation and execute these trajectories on the robot. This is to study whether the considered task is complex enough.
    \item \textbf{CT-Sensor.} We train a policy that uses a continuous-valued sensor input rather than the binarized version. This is to study if using continuous signals will lead to Sim2Real difficulty.
\end{enumerate}
 Note that there also exist some vision-based dexterous manipulation baselines like~\cite{chen2022visual, dextreme}. However, we do not compare our method to theirs as they require a collection of a very large amount of visual simulation data, which takes significantly longer real-world time.

\subsection{Sim: Policy learning with different sensing capabilities.}
In this section, we study whether our tactile policy and the considered baseline policies are able to succeed in the training environments in the simulation. We study both the single and multi objects setup. We use the cuboid as the object in single-object training, which is common in the previous works~\cite{rl-openai, dextreme}. We use object set A for multi-object training. The results are shown in Figure~\ref{fig:curve}. We find that in the single object setup, both the No-Sense and LS-Sensor policies have a lower rotation reward compared with our policy. Interestingly, we find that LS-Sensor can achieve a higher duration~(TTF) compared with No-Sensor and can match that of our full system. This result suggests that tactile sensors of low sensitivity may still be useful to make the motion more secure. For the multi-object training, we find that our tactile policy outperforms the baseline policies by a large margin. The baseline policies fail completely in this case, while our policy is still able to succeed. This result indicates that using tactile information is essential to tame touch-only multi-object rotation. The failure of the LS-Sensor in the multi-object training case suggests that having a high-sensitivity sensor to sense the slightest contact is important.

\subsection{Sim: Is a tactile policy robust and generalizable?}
Though our tactile policy and the baseline policies can succeed in some cases during training, so far it remains unknown whether they are robust and generalizable. We consider a policy robust if it can perform well on the unseen physics parameter setup on the same set of objects. We consider a policy generalizable if it can perform well on an unseen set of objects. 

We test the robustness of the single-object setting. To do this, we sample from a smaller, unseen range of friction and mass parameters, and perform rollout. In this case, the object is more likely to slide in the hand, requiring the hand to manipulate it in a more careful manner. The results are shown in Table~\ref{table:abl1}. We find that there is little performance drop in our full method. However, for No-Sensor and DS-Sensor, we can observe a clear performance drop. We also find that the low-sensitivity policy LS-Sensor also performs well in the unseen physics setup. This suggests that a low-sensitivity tactile sensing ability is sufficient for the robustness of the single object rotation.

The generalization testing result is shown in Table~\ref{table:abl2}. We train the policies on object set A and test them on object set B. Since No-Sensor and LS-Sensor baseline does not work on the multi-object training setup, we only compare our method with DS-Sensor. We find that disabling the sensor input will lead to a significant performance drop both in the seen and unseen object setup. This result suggests that tactile information is indeed important for generalization.

\begin{table*}

\centering
\captionof{table}{Performance of different methods on the multi-object rotation task on the real robot. The results are averaged on 3 policies trained on 3 seeds. Each trial lasts 30 seconds. The CRA metric is measured by the number of turned rounds. The TTF metric is measured in seconds. Our proposed method can rotate both the seen and unseen objects. } 
\normalsize
\renewcommand\arraystretch{1.1}
\setlength\tabcolsep{3pt}
 \begin{tabular*}{\textwidth}{l@{\extracolsep{\fill}}cccccccccc}
\toprule
\textbf{Seen} &\multicolumn{2}{c}{\textbf{Object C1}} & \multicolumn{2}{c}{\textbf{Object C2}} & \multicolumn{2}{c}{\textbf{Object C3}} & \multicolumn{2}{c}{\textbf{Object C4}} & 
\multicolumn{2}{c}{\textbf{Object C5}}  \\
& CRA & TTF    & CRA & TTF  & CRA & TTF   & CRA & TTF   & CRA & TTF    \\
\hline
OL & \ms{0.58}{0.14} & \ms{13.30}{7.77} & \ms{0.08}{0.14} & \ms{4.67}{8.08}  & \ms{0.75}{0.66} & \ms {18.67}{16.29}  & \ms{0.50}{0} & \ms{24.00}{5.29}  & \ms{0.83}{1.04} & \ms {13.67}{15.18}    \\
No-Sensor{ }& \ms{0.25}{0.25}  &  \ms{7.67}{6.80} & \ms{0.33}{0.28} & \ms{14.7}{15.01}  & \ms{0.08}{0.144} & \ms{3.67}{6.35}  & \ms{0.42}{0.14} & \ms{16.00}{12.17}  & \ms{0.25}{0.25} & \ms{12.67}{15.53}   \\
CT-Sensor{ } & \ms{2.50}{3.25} & \ms{20.00}{8.66} & \ms{0.75}{0.66} & \ms{17.67}{10.79}  & \ms{2.42}{2.10} & \ms{15.33}{15.01}  & \ms{1.92}{1.46} & \ms{23.00}{12.12}  & \ms{1.00}{0.87} & \ms{17.00}{15.39}  \\
Ours & \ms{\textbf{4.91}}{\textbf{0.52}}  & \ms{\textbf{30.00}}{\textbf{0.00}}& \ms{\textbf{2.83}}{\textbf{1.26}} & \ms{\textbf{28.67}}{\textbf{2.31}}  & \ms{\textbf{2.92}}{\textbf{1.38}} & \ms{\textbf{30.00}}{\textbf{0 .00}}  & \ms{\textbf{4.50}}{\textbf{1.73}} & \ms{\textbf{30.00}}{\textbf{0.00}}  & \ms{\textbf{2.00}}{\textbf{0.00}} & \ms{\textbf{26.67}}{\textbf{5.77}}   \\
\midrule
\textbf{Unseen} &\multicolumn{2}{c}{\textbf{Tomato}} & \multicolumn{2}{c}{\textbf{Apple}} & \multicolumn{2}{c}{\textbf{Orange}} & \multicolumn{2}{c}{\textbf{Soupcan}} & 
\multicolumn{2}{c}{\textbf{Rubber Duck}}  \\
& CRA & TTF    & CRA & TTF  & CRA & TTF   & CRA & TTF   & CRA & TTF    \\
\hline
OL & \ms{0.25}{0.25} & \ms{20.00}{17.32} & \ms{0.67}{0.76} & \ms{20.00}{17.32}  & \ms{0.5}{0.87} & \ms{10.00}{17.32}  & \ms{1.5}{1.32} & \ms{20.00}{17.32}  & \ms{0.33}{0.29} & \ms{20.00}{17.32}  \\
No-Sensor & \ms{0.00}{0.00} & \ms{0.00}{0.00} & \ms{0.33}{0.58} & \ms{10.00}{17.32}  & \ms{0.75}{1.09} & \ms{12.33}{15.70}  & \ms{0.08}{0.14} & \ms{2.00}{3.46}  & \ms{0.33}{0.29} & \ms{20.00}{17.32}  \\
CT-Sensor & \ms{0.33}{0.29} & \ms{12.33}{15.70} & \ms{0.42}{0.52} & \ms{15.33}{15.01}  & \ms{2.08}{2.10} & \ms{24.33}{4.93}  & \ms{2.08}{2.79} & \ms{19.33}{16.77}  & \ms{\textbf{1.50}}{\textbf{0.75}} & \ms{\textbf{30.00}}{\textbf{0.00}}  \\
Ours & \ms{\textbf{1.08}}{\textbf{0.14}} & \ms{\textbf{27.33}}{\textbf{4.62}} & \ms{\textbf{2.67}}{\textbf{1.04}} & \ms{\textbf{30.00}}{\textbf{0.00}}  & \ms{\textbf{3.00}}{\textbf{1.32}} & \ms{\textbf{30.00}}{\textbf{0.00}}  & \ms{\textbf{4.25}}{\textbf{1.56}} & \ms{\textbf{27.33}}{\textbf{4.62}}  & \ms{1.42}{0.38} & \ms{29.00}{1.73} \\
\bottomrule
\end{tabular*}
\label{table:real}
\end{table*}

\begin{table}[t]
\captionof{table}{Performance of different methods on the single-object rotation task with physics distribution shift in simulation. The results are averaged on 3 seeds.} 
\vspace{0.1cm}
\normalsize
\renewcommand\arraystretch{1.05}
\begin{tabular}{lcccc}
\toprule
Method &\multicolumn{2}{c}{Seen Physics Setup} & \multicolumn{2}{c}{Unseen Physics Setup} \\
\cline{2-3}  \cline{4-5} 
& CRR & TTF    & CRR & TTF \\
\hline
No-Sensor & \ms{689.3}{141.5} & \ms{33.3}{4.7} & \ms{369.0}{129.1} & \ms{23.5}{6.1} \\
Sensor & \msb{963.8}{377.8} & \msb{42.2}{4.1} & \msb{919.3}{338.0} & \msb{40.0}{4.3} \\
DS-Sensor & \ms{904.2}{408.6} & \ms{39.1}{6.3} & \ms{615.5}{293.2} & \ms{31.2}{8.0} \\
LS-Sensor & \ms{860.0}{348.7} & \ms{38.8}{6.9} & \ms{796.5}{366.7} & \ms{37.4}{8.4} \\
\bottomrule
\end{tabular}
\label{table:abl1}
\end{table}

\begin{table}[t]
\vspace{-0.2cm}
\captionof{table}{Performance of different methods on the multi-object rotation task in simulation. The results are averaged on 3 seeds.} 
\vspace{0.1cm}
\normalsize
\renewcommand\arraystretch{1.05}
\begin{tabular}{lcccc}
\toprule
Method &\multicolumn{2}{c}{Seen Object Setup} & \multicolumn{2}{c}{Unseen Object Setup} \\
\cline{2-3}  \cline{4-5} 
& CRR & TTF    & CRR & TTF \\
\hline
Sensor & \msb{976.1}{86.5} & \msb{42.1}{0.6} & \msb{594.4}{63.2} & \msb{28.2}{2.7} \\
DS-Sensor & \ms{351.5}{28.0} & \ms{18.6}{0.7} & \ms{186.5}{16.1} & \ms{10.7}{1.4} \\
\bottomrule
\vspace{-0.2cm}
\end{tabular}
\label{table:abl2}
\vspace{-0.2cm}
\end{table}

\subsection{Real: Dexterity without Vision}

We have seen that our tactile policy can achieve superior performance in the ideal simulation setup. Now, we transfer the trained tactile policy to the real robot and verify if it still offers the demonstrated benefit. In real-world experiments, we train the baselines on object sets A and B and evaluate these policies on object set C. We evaluate each method using 3 different seeds for each object. The results are shown in Table~\ref{table:real}. 

Our method can outperform all the baselines. It can not only perform rotation on the seen, artificial training objects but also generalize to unseen real-world objects like apples and tomatoes. The method with continuous tactile sensor signals also performs better than the other two methods without feedback. We observe that both the no-sensor and the open-loop policy can at most rotate the object for 180 degrees on the evaluated objects, after which they will get stuck or push the object off the palm, resulting in a failure. In addition, by studying the behavior of our policy and baselines, we find that our policy can adjust the finger motion immediately when objects get to positions that are easy to get stuck or fall. In contrast, the baselines without sensors do not have such kind of adaptive behavior. This result suggests that it is crucial to have a dose of tactile feedback in the considered in-hand object manipulation setup. By comparing the methods with continuous contact signals and binarized contact signals, we find that the latter has better performance. Even though the policy with continuous signals can perform well in some objects, it has poor generalizability and huge variance between different objects. This may be due to the huge gap in force measurement between simulation and the real world. 

\begin{figure*}[ht]
\centering
\includegraphics[width=\textwidth]{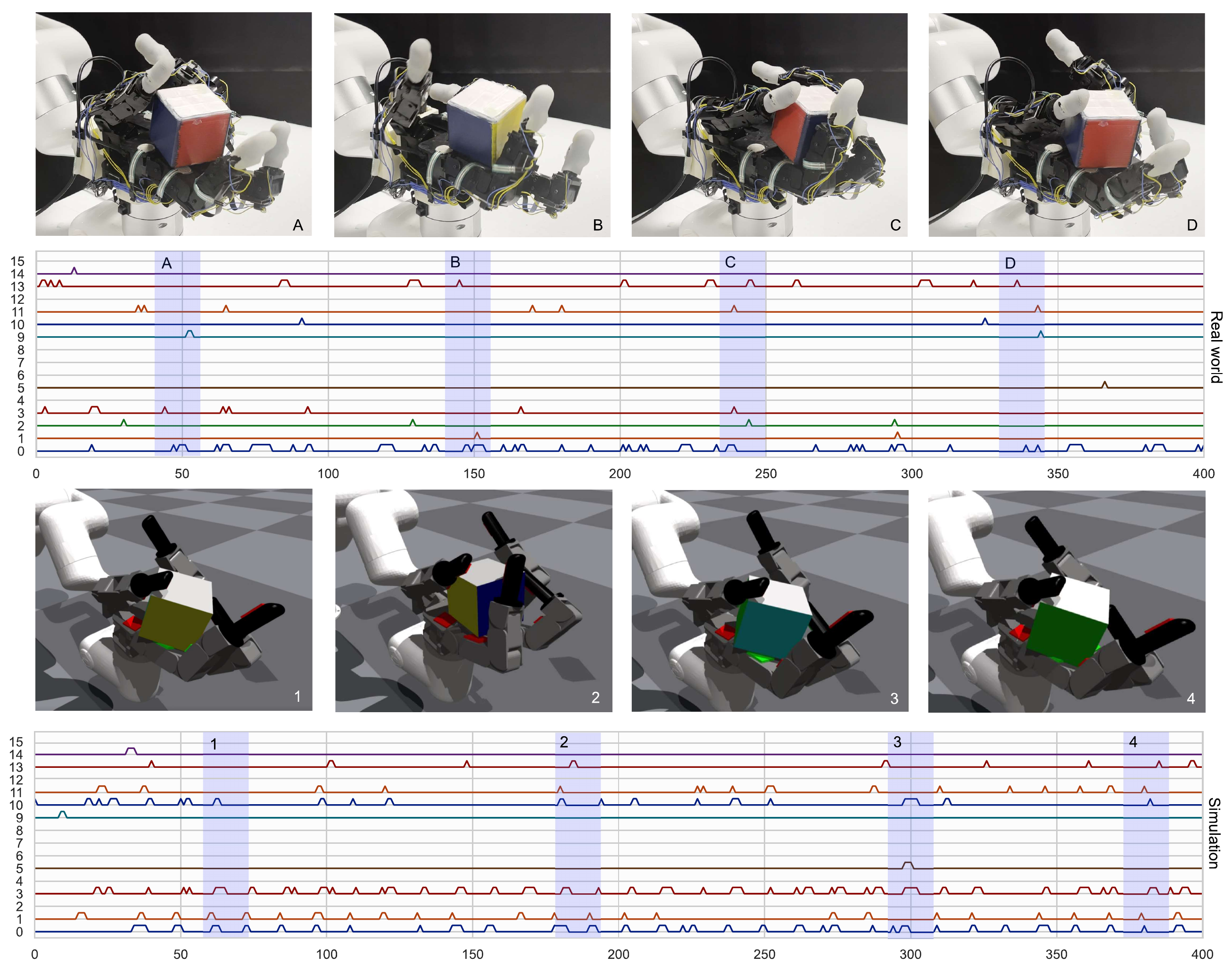}
 \caption{Visualization of contact signals in 400 steps in the simulation and real-world experiments on a cuboid. We also show some typical frames during the rotation process. We can see that the contact signals in the simulation and the real world in general align. This accounts for successful Sim2Real transfer.}
\label{fig:response}
\vspace{-0.65cm}
\end{figure*}

\begin{figure*}[ht]
\centering
\includegraphics[width=\textwidth]{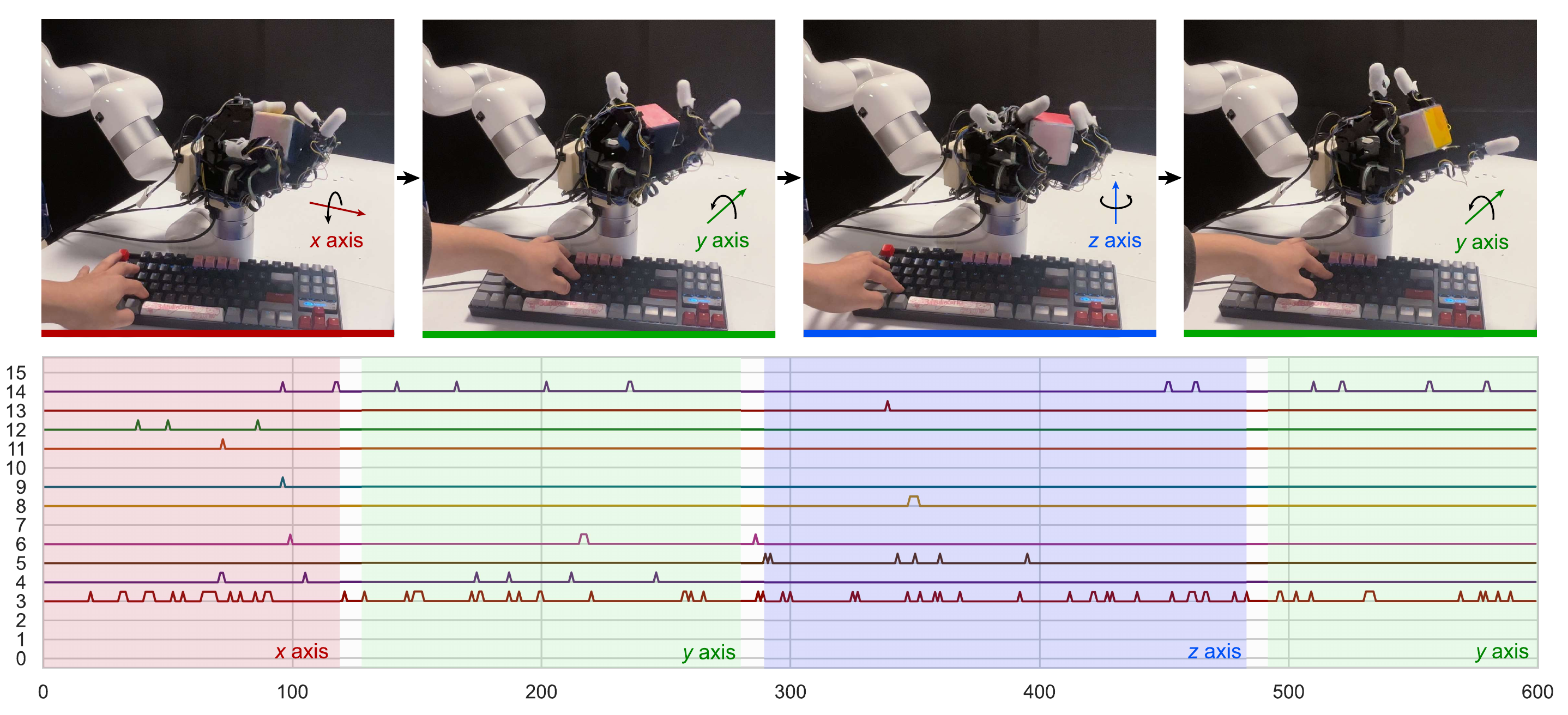}
 \caption{With the learned rotation primitives around $x$, $y$, and $z$ axis, we can perform human-robot shared control to reorient an object. In this example, a human operator uses a keyboard to rotate a cuboid around $x, y, z, y$ axes consecutively. We also visualize the contact signal throughout this 600-step process~(60 seconds). }
\label{fig:human}
\vspace{-0.4cm}
\end{figure*}

\subsection{Qualitative Analysis: Sensor Response}
To understand why Sim2Real can be successful, we conduct a case study on cuboid rotation to analyze the sensor response. We visualize two 40 seconds test trajectories recorded in the simulation and real in Figure~\ref{fig:response} during the test. It is worthwhile mentioning that different runs in real will produce different patterns, and we put more cases in the appendix. We find that the contact signals in the simulation are slightly denser (along the temporal x-axis) and richer (along the sensor y-axis) compared with that of the real. Some sensors are also more likely to be activated~(e.g., sensors 1 and 10) in the simulation, but the overall patterns of the simulation and the real are similar. This can explain why our Sim2Real transfer is successful. Moreover, when looking at local windows used by our policy~(0.4s) in simulation, we observe that there are various, diverse sensor activation patterns. We hypothesize that learning from such a diverse distribution could also help the policy to transfer to the sensor observation in the real world. 

\subsection{Ablation Study I: Importance Analysis of Sensors}
Then, we perform ablation studies of the system on the real robot to see which sensors are more important for a successful rotation. We divide the sensors into two groups: Fingertip and Palm. We disable these two groups of sensors and train two policies (No-Fingertip and No-Palm). Then we compare them to our full policy and DS-Sensor, see Table~\ref{table:abl_real}. We find that neither of the two considered policies can compare to our full policy. They achieve a similar performance as DS-Sensor, which suggests that both groups of sensors are essential for the success of in-hand object rotation.
\begin{figure}[ht]
\subfigure{\includegraphics[width=1.0\linewidth]{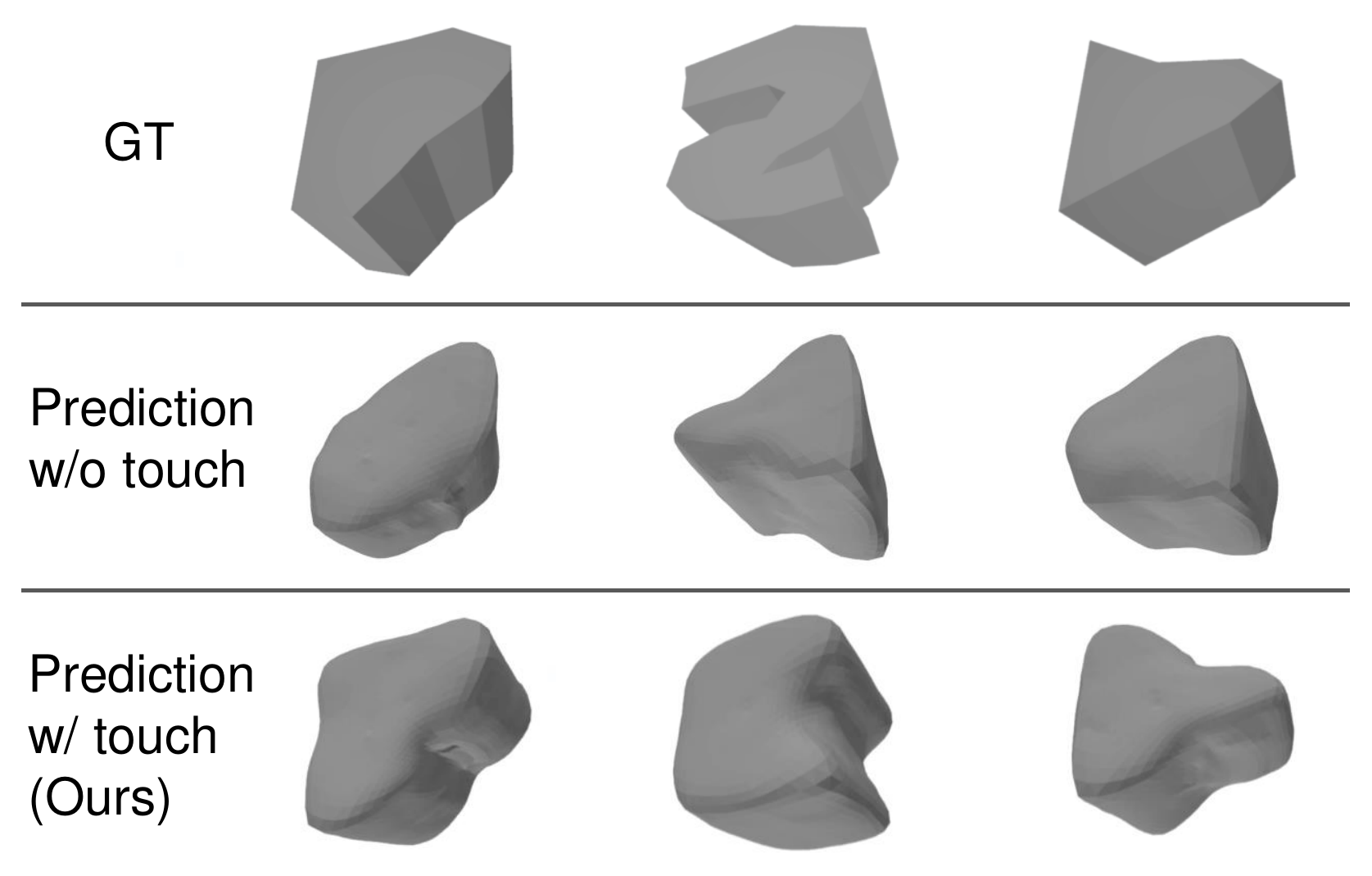}}

\caption{Qualitative mesh reconstruction results in simulation. When the touch information does not present, we can not infer the shape of the rotated object accurately. In contrast, our method is able to reconstruct the groundtruth object by a 20-second rotation.}
\label{fig:recon}
\vspace{-0.3cm}
\end{figure}

\begin{table}[t]
\captionof{table}{Ablation analysis of the system. We train policies on different sensor setups and test their performance. The results are averaged on 3 seeds.} 
\vspace{0.1cm}
\normalsize
\renewcommand\arraystretch{1.05}
\setlength\tabcolsep{3.4pt}
\begin{tabular}{lcccc}
\toprule
Method &\multicolumn{2}{c}{Cuboid} & \multicolumn{2}{c}{Rubber Duck} \\
\cline{2-3}  \cline{4-5} 
& CRR & TTF    & CRR & TTF \\
\hline
Sensor & \ms{\textbf{4.91}}{\textbf{0.52}}  & \ms{\textbf{30.00}}{\textbf{0.00}} & \ms{\textbf{1.42}}{\textbf{0.38}} & \ms{\textbf{29.00}}{\textbf{1.73}} \\
DS-Sensor & \ms{0.25}{0.25}  &  \ms{7.67}{6.80} & \ms{0.33}{0.29} & \ms{20.00}{17.32}  \\
No-Fingertip & \ms{0.17}{0.29} & \ms{3.33}{5.77} & \ms{0.42}{0.14} & \ms{17.00}{2.64} \\
No-Palm & \ms{0.42}{0.38} & \ms{17.00}{14.73} & \ms{0.42}{0.14} & \ms{16.67}{11.72} \\
\bottomrule
\end{tabular}
\label{table:abl_real}
\vspace{-0.4cm}
\end{table}

\subsection{Ablation Study II: A Shape Understanding Perspective}
So far, we have seen that the tactile information is essential for successful object rotation. In this part, our goal is to understand its success from a shape understanding perspective. We study whether our tactile information can reveal the shape information of the object, which may be helpful for learning robust and adaptive rotation behavior across different objects. Specifically, we would like to see if it is possible to predict the shape of the object using the rollout of a rotation policy. For simplicity, we focus on the $z$-axis rotation of column-shaped objects. We first train a $z$-axis rotation policy on 125 different irregular, column-shaped objects. Then, we use this policy to collect 55000 policy rollouts of rotating these objects, and each of these rollouts lasts 200 control steps~(20 seconds). Next, we split these collected rollouts into a training dataset and a test dataset. The objects in the test dataset do not present in the training dataset. We train a temporal-CNN model to predict the shape of the object using the full rollout trajectory as input, and then we use the trained model to reconstruct the shape of objects in the test dataset. We compare our model to another ablated model, which discards all the tactile observation in the rollout during prediction~(by setting them to 0). The shape reconstruction mean squared error~(MSE) of our model is 0.22, while that of the ablated model is 0.45. This suggest that using tactile infomation can indeed help shape understanding. Moreover, we provide visualization of predicted object shapes are in Figure~\ref{fig:recon}. With the tactile sensors, our model can reconstruct object shape much better than the ablated model. The shape understanding results suggest that the binarized tactile information is indeed important for the robot to percept the object and interact with it in a meaningful way.

\subsection{Rotation Around Other Axes}
Besides the rotation around the $z$ axis, we also test whether our system is able to perform rotation around other axes. Here, we study the rotation around the $x$ and $y$ axes. To do so, we train our policy on the object set A and B as in the previous experiments. The results are shown in Table~\ref{table:summ_all}. We find that our system is still able to rotate most of the objects successfully, though it may have difficulty rotation some particular objects which results in lower CRAs. We observe that rotation around $x$ and $y$ axes involves many critical contacts between the object and the side of finger links. This may explain why failures can occur since the layout of our current sensor array does not support this feature. We hypothesis that a denser contact sensor array over each finger link can remedy this problem. 

The rotation around $x$, $y$, and $z$ axis provides a useful set of primitives. This enables human to use high-level commands to control the rotation behavior, as shown in Figure~\ref{fig:human}. In this example, A human operator presses the keyboard to send different rotation commands (i.e. around $x$, $y$, or $z$). The robot hand is then able to execute the desired rotation, reorienting the object to different poses. 

\begin{table}[t]
\captionof{table}{Summary of rotation performance around different axes.  We provide the averaged results over the object set on the $x$, $y$, and $z$ axis rotations. The results are averaged on 3 seeds.} 
\vspace{0.1cm}
\normalsize
\renewcommand\arraystretch{1.05}
\centering
\begin{tabular}{lcccc}
\toprule
Rotation &\multicolumn{2}{c}{Seen Obj} & \multicolumn{2}{c}{Unseen Obj} \\
\cline{2-3}  \cline{4-5} 
& CRR & TTF    & CRR & TTF \\
\hline
$x$-axis & \ms{1.68}{0.78}  & \ms{24.13}{6.04} & \ms{2.71}{1.37} & \ms{18.2}{9.19} \\
$y$-axis & \ms{1.88}{0.38}  &  \ms{22.46}{4.81} & \ms{1.05}{0.56} & \ms{23.13}{3.01}  \\
$z$-axis & \ms{3.43}{1.22} & \ms{29.06}{1.45} & \ms{2.48}{1.27} & \ms{28.73}{1.34} \\
\bottomrule
\end{tabular}
\label{table:summ_all}
\vspace{-0.4cm}
\end{table}

 \section{Conclusion}
 In this paper, we have presented Touch Dexterity, a new dexterous manipulation system that is able to rotate different objects through touch without vision. We showed an end-to-end reinforcement learning framework to learn dexterous manipulation skills on the proposed system. We carried out experiments both in simulation and real to demonstrate its effectiveness. Our work demonstrated that we are able to achieve touch-only dexterity as humans in real for the first time. In the future, there are many promising future directions to investigate, such as exploring the use of a more dense contact sensor array and scaling up the system to solve more diverse tasks. We hope that our work can pave the way for more intelligent robot hands.

\bibliographystyle{plainnat}
\bibliography{references}

\begin{thebibliography}{76}
\providecommand{\natexlab}[1]{#1}
\providecommand{\url}[1]{\texttt{#1}}
\expandafter\ifx\csname urlstyle\endcsname\relax
  \providecommand{\doi}[1]{doi: #1}\else
  \providecommand{\doi}{doi: \begingroup \urlstyle{rm}\Url}\fi

\bibitem[Aiyama et~al.(1993)Aiyama, Inaba, and Inoue]{pivot_dex}
Y.~Aiyama, M.~Inaba, and H.~Inoue.
\newblock Pivoting: A new method of graspless manipulation of object by robot
  fingers.
\newblock In \emph{IEEE/RSJ International Conference on Intelligent Robots and
  Systems (IROS)}, 1993.

\bibitem[Andrychowicz et~al.(2020)Andrychowicz, Baker, Chociej, Jozefowicz,
  McGrew, Pachocki, Petron, Plappert, Powell, Ray, et~al.]{rl-openai}
OpenAI:~Marcin Andrychowicz, Bowen Baker, Maciek Chociej, Rafal Jozefowicz, Bob
  McGrew, Jakub Pachocki, Arthur Petron, Matthias Plappert, Glenn Powell, Alex
  Ray, et~al.
\newblock Learning dexterous in-hand manipulation.
\newblock \emph{The International Journal of Robotics Research~(IJRR)},
  39\penalty0 (1):\penalty0 3--20, 2020.

\bibitem[Arunachalam et~al.(2022{\natexlab{a}})Arunachalam, G{\"u}zey,
  Chintala, and Pinto]{arunachalam2022holo}
Sridhar~Pandian Arunachalam, Irmak G{\"u}zey, Soumith Chintala, and Lerrel
  Pinto.
\newblock Holo-dex: Teaching dexterity with immersive mixed reality.
\newblock \emph{arXiv preprint arXiv:2210.06463}, 2022{\natexlab{a}}.

\bibitem[Arunachalam et~al.(2022{\natexlab{b}})Arunachalam, Silwal, Evans, and
  Pinto]{arunachalam2022dexterous}
Sridhar~Pandian Arunachalam, Sneha Silwal, Ben Evans, and Lerrel Pinto.
\newblock Dexterous imitation made easy: A learning-based framework for
  efficient dexterous manipulation.
\newblock \emph{arXiv preprint arXiv:2203.13251}, 2022{\natexlab{b}}.

\bibitem[Bai and Liu(2014)]{dexhand_3_bai2014dexterous}
Yunfei Bai and C.~Karen Liu.
\newblock Dexterous manipulation using both palm and fingers.
\newblock In \emph{2014 IEEE International Conference on Robotics and
  Automation (ICRA)}, pages 1560--1565, 2014.
\newblock \doi{10.1109/ICRA.2014.6907059}.

\bibitem[Bhatt et~al.(2021)Bhatt, Sieler, Puhlmann, and
  Brock]{dexhand_2_Bhatt2021surprisingly}
Aditya Bhatt, Adrian Sieler, Steffen Puhlmann, and Oliver Brock.
\newblock Surprisingly robust in-hand manipulation: An empirical study.
\newblock In \emph{Robotics: Science and Systems~(RSS)}, 2021.

\bibitem[Bhattacharjee et~al.(2015)Bhattacharjee, Wade, and
  Kemp]{bhattacharjee2015material}
Tapomayukh Bhattacharjee, Joshua Wade, and Charles~C Kemp.
\newblock Material recognition from heat transfer given varying initial
  conditions and short-duration contact.
\newblock In \emph{Robotics: Science and Systems~(RSS)}, 2015.

\bibitem[Bhirangi et~al.(2021)Bhirangi, Hellebrekers, Majidi, and
  Gupta]{bhirangi2021reskin}
Raunaq Bhirangi, Tess Hellebrekers, Carmel Majidi, and Abhinav Gupta.
\newblock Reskin:versatile, replaceable, lasting tactile skins.
\newblock In \emph{Conference on Robot Learning~(CoRL)}, 2021.

\bibitem[Bi et~al.(2021)Bi, Sferrazza, and D’Andrea]{bi2021zero}
Thomas Bi, Carmelo Sferrazza, and Raffaello D’Andrea.
\newblock Zero-shot sim-to-real transfer of tactile control policies for
  aggressive swing-up manipulation.
\newblock \emph{IEEE Robotics and Automation Letters}, 6\penalty0 (3):\penalty0
  5761--5768, 2021.

\bibitem[Bicchi and Sorrentino(1995)]{dex_4}
A.~Bicchi and R.~Sorrentino.
\newblock Dexterous manipulation through rolling.
\newblock In \emph{Proceedings of 1995 IEEE International Conference on
  Robotics and Automation}, volume~1, pages 452--457 vol.1, 1995.
\newblock \doi{10.1109/ROBOT.1995.525325}.

\bibitem[Buescher et~al.(2015)Buescher, Meier, Walck, Haschke, and
  Ritter]{tactile_skin}
Gereon Buescher, Martin Meier, Guillaume Walck, Robert Haschke, and Helge~J
  Ritter.
\newblock Augmenting curved robot surfaces with soft tactile skin.
\newblock In \emph{2015 IEEE/RSJ International Conference on Intelligent Robots
  and Systems (IROS)}, 2015.

\bibitem[Chavan-Dafle and Rodriguez(2020)]{dex_7}
Nikhil Chavan-Dafle and Alberto Rodriguez.
\newblock Sampling-based planning of in-hand manipulation with external pushes.
\newblock In \emph{Robotics Research: The 18th International Symposium ISRR},
  pages 523--539. Springer, 2020.

\bibitem[Chebotar et~al.(2014)Chebotar, Kroemer, and Peters]{6943031}
Yevgen Chebotar, Oliver Kroemer, and Jan Peters.
\newblock Learning robot tactile sensing for object manipulation.
\newblock In \emph{IEEE/RSJ International Conference on Intelligent Robots and
  Systems~(IROS)}, 2014.

\bibitem[Chen et~al.(2022{\natexlab{a}})Chen, Tippur, Wu, Kumar, Adelson, and
  Agrawal]{chen2022visual}
Tao Chen, Megha Tippur, Siyang Wu, Vikash Kumar, Edward Adelson, and Pulkit
  Agrawal.
\newblock Visual dexterity: In-hand dexterous manipulation from depth.
\newblock \emph{arXiv preprint arXiv:2211.11744}, 2022{\natexlab{a}}.

\bibitem[Chen et~al.(2022{\natexlab{b}})Chen, Xu, and Agrawal]{corl21-inhand}
Tao Chen, Jie Xu, and Pulkit Agrawal.
\newblock A system for general in-hand object re-orientation.
\newblock In \emph{Conference on Robot Learning~(CoRL)}, pages 297--307,
  2022{\natexlab{b}}.

\bibitem[Cherif and Gupta(1999{\natexlab{a}})]{dex_1}
Mo{\"e}z Cherif and Kamal~K Gupta.
\newblock Planning quasi-static fingertip manipulations for reconfiguring
  objects.
\newblock \emph{IEEE Transactions on Robotics and Automation}, 15\penalty0
  (5):\penalty0 837--848, 1999{\natexlab{a}}.

\bibitem[Cherif and Gupta(1999{\natexlab{b}})]{sliding_dex}
Mo{\"e}z Cherif and Kamal~K Gupta.
\newblock Planning quasi-static fingertip manipulations for reconfiguring
  objects.
\newblock \emph{IEEE Transactions on Robotics and Automation}, 15\penalty0
  (5):\penalty0 837--848, 1999{\natexlab{b}}.

\bibitem[Church et~al.(2022)Church, Lloyd, Lepora, et~al.]{church2022tactile}
Alex Church, John Lloyd, Nathan~F Lepora, et~al.
\newblock Tactile sim-to-real policy transfer via real-to-sim image
  translation.
\newblock In \emph{Conference on Robot Learning~(CoRL)}, 2022.

\bibitem[Dong et~al.(2021)Dong, Jha, Romeres, Kim, Nikovski, and
  Rodriguez]{tactile_insertion}
Siyuan Dong, Devesh~K Jha, Diego Romeres, Sangwoon Kim, Daniel Nikovski, and
  Alberto Rodriguez.
\newblock Tactile-rl for insertion: Generalization to objects of unknown
  geometry.
\newblock In \emph{IEEE International Conference on Robotics and Automation
  (ICRA)}, 2021.

\bibitem[Doshi et~al.(2022)Doshi, Taylor, and Rodriguez]{contact_ar}
Neel Doshi, Orion Taylor, and Alberto Rodriguez.
\newblock Manipulation of unknown objects via contact configuration regulation.
\newblock In \emph{International Conference on Robotics and Automation (ICRA)},
  2022.

\bibitem[Doulgeri and Droukas(2013)]{dex_5}
Zoe Doulgeri and Leonidas Droukas.
\newblock On rolling contact motion by robotic fingers via prescribed
  performance control.
\newblock In \emph{IEEE International Conference on Robotics and
  Automation~(ICRA)}, 2013.

\bibitem[Driess et~al.(2017)Driess, Englert, and Toussaint]{bin_contact_2}
Danny Driess, Peter Englert, and Marc Toussaint.
\newblock Active learning with query paths for tactile object shape
  exploration.
\newblock In \emph{IEEE/RSJ International Conference on Intelligent Robots and
  Systems (IROS)}, 2017.

\bibitem[Gao et~al.(2022)Gao, Si, Chang, Clarke, Bohg, Fei-Fei, Yuan, and
  Wu]{gao2022objectfolder}
Ruohan Gao, Zilin Si, Yen-Yu Chang, Samuel Clarke, Jeannette Bohg, Li~Fei-Fei,
  Wenzhen Yuan, and Jiajun Wu.
\newblock Objectfolder 2.0: A multisensory object dataset for sim2real
  transfer.
\newblock In \emph{Proceedings of the IEEE/CVF Conference on Computer Vision
  and Pattern Recognition}, pages 10598--10608, 2022.

\bibitem[Habib et~al.(2014)Habib, Ranatunga, Shook, and Popa]{habib2014skinsim}
Ahsan Habib, Isura Ranatunga, Kyle Shook, and Dan~O Popa.
\newblock Skinsim: A simulation environment for multimodal robot skin.
\newblock In \emph{IEEE International Conference on Automation Science and
  Engineering (CASE)}, 2014.

\bibitem[Han and Trinkle(1998)]{dex_3}
L.~Han and J.C. Trinkle.
\newblock Dextrous manipulation by rolling and finger gaiting.
\newblock In \emph{IEEE International Conference on Robotics and Automation
  (ICRA)}, 1998.

\bibitem[Han et~al.(1997{\natexlab{a}})Han, Guan, Li, Shi, and Trinkle]{dex_2}
Li~Han, Yi-Sheng Guan, ZX~Li, Q~Shi, and Jeffrey~C Trinkle.
\newblock Dextrous manipulation with rolling contacts.
\newblock In \emph{IEEE International Conference on Robotics and
  Automation~(ICRA)}, 1997{\natexlab{a}}.

\bibitem[Han et~al.(1997{\natexlab{b}})Han, Guan, Li, Shi, and
  Trinkle]{rolling_dex}
Li~Han, Yi-Sheng Guan, ZX~Li, Q~Shi, and Jeffrey~C Trinkle.
\newblock Dextrous manipulation with rolling contacts.
\newblock In \emph{International Conference on Robotics and Automation~(ICRA)},
  1997{\natexlab{b}}.

\bibitem[Handa et~al.(2023)Handa, Allshire, Makoviychuk, Petrenko, Singh, Liu,
  Makoviichuk, Van~Wyk, Zhurkevich, Sundaralingam, et~al.]{dextreme}
Ankur Handa, Arthur Allshire, Viktor Makoviychuk, Aleksei Petrenko, Ritvik
  Singh, Jingzhou Liu, Denys Makoviichuk, Karl Van~Wyk, Alexander Zhurkevich,
  Balakumar Sundaralingam, et~al.
\newblock Dextreme: Transfer of agile in-hand manipulation from simulation to
  reality.
\newblock In \emph{International Conference on Robotics and Automation (ICRA)},
  2023.

\bibitem[Hogan et~al.(2020)Hogan, Ballester, Dong, and
  Rodriguez]{tactile2020alberto}
Francois~R Hogan, Jose Ballester, Siyuan Dong, and Alberto Rodriguez.
\newblock Tactile dexterity: Manipulation primitives with tactile feedback.
\newblock In \emph{IEEE International Conference on Robotics and Automation
  (ICRA)}, 2020.

\bibitem[Johansson and Westling(1984)]{human_tactile}
Roland~S Johansson and Goran Westling.
\newblock Roles of glabrous skin receptors and sensorimotor memory in automatic
  control of precision grip when lifting rougher or more slippery objects.
\newblock \emph{Experimental brain research}, 56:\penalty0 550--564, 1984.

\bibitem[Kolamuri et~al.(2021)Kolamuri, Si, Zhang, Agarwal, and
  Yuan]{kolamuri2021improving_grasp}
Raj Kolamuri, Zilin Si, Yufan Zhang, Arpit Agarwal, and Wenzhen Yuan.
\newblock Improving grasp stability with rotation measurement from tactile
  sensing.
\newblock In \emph{IEEE/RSJ International Conference on Intelligent Robots and
  Systems (IROS)}, 2021.

\bibitem[Kumar et~al.(2014)Kumar, Tassa, Erez, and
  Todorov]{dexhand_1_kumar2014real_time_behaviour_synthesis}
Vikash Kumar, Yuval Tassa, Tom Erez, and Emanuel Todorov.
\newblock Real-time behaviour synthesis for dynamic hand-manipulation.
\newblock In \emph{2014 IEEE International Conference on Robotics and
  Automation (ICRA)}, pages 6808--6815. IEEE, 2014.

\bibitem[Lambeta et~al.(2020)Lambeta, Chou, Tian, Yang, Maloon, Most, Stroud,
  Santos, Byagowi, Kammerer, et~al.]{lambeta2020digit}
Mike Lambeta, Po-Wei Chou, Stephen Tian, Brian Yang, Benjamin Maloon,
  Victoria~Rose Most, Dave Stroud, Raymond Santos, Ahmad Byagowi, Gregg
  Kammerer, et~al.
\newblock Digit: A novel design for a low-cost compact high-resolution tactile
  sensor with application to in-hand manipulation.
\newblock \emph{IEEE Robotics and Automation Letters}, 5\penalty0 (3):\penalty0
  3838--3845, 2020.

\bibitem[Lee et~al.(2019)Lee, Zhu, Zachares, Tan, Srinivasan, Savarese,
  Fei-Fei, Garg, and Bohg]{https://doi.org/10.48550/arxiv.1907.13098}
Michelle~A. Lee, Yuke Zhu, Peter Zachares, Matthew Tan, Krishnan Srinivasan,
  Silvio Savarese, Li~Fei-Fei, Animesh Garg, and Jeannette Bohg.
\newblock Making sense of vision and touch: Learning multimodal representations
  for contact-rich tasks, 2019.
\newblock URL \url{https://arxiv.org/abs/1907.13098}.

\bibitem[Li et~al.(2020)Li, Kroemer, Su, Veiga, Kaboli, and
  Ritter]{tactile2020review}
Qiang Li, Oliver Kroemer, Zhe Su, Filipe~Fernandes Veiga, Mohsen Kaboli, and
  Helge~Joachim Ritter.
\newblock A review of tactile information: Perception and action through touch.
\newblock \emph{IEEE Transactions on Robotics}, 36\penalty0 (6):\penalty0
  1619--1634, 2020.

\bibitem[Liang et~al.(2020)Liang, Handa, Van~Wyk, Makoviychuk, Kroemer, and
  Fox]{liang2020hand}
Jacky Liang, Ankur Handa, Karl Van~Wyk, Viktor Makoviychuk, Oliver Kroemer, and
  Dieter Fox.
\newblock In-hand object pose tracking via contact feedback and gpu-accelerated
  robotic simulation.
\newblock In \emph{IEEE International Conference on Robotics and Automation
  (ICRA)}, pages 6203--6209. IEEE, 2020.

\bibitem[Liu et~al.(2022)Liu, Pathak, and Kitani]{liu2022herd}
Xingyu Liu, Deepak Pathak, and Kris~M Kitani.
\newblock Herd: Continuous human-to-robot evolution for learning from human
  demonstration.
\newblock \emph{arXiv preprint arXiv:2212.04359}, 2022.

\bibitem[Makoviychuk et~al.(2021)Makoviychuk, Wawrzyniak, Guo, Lu, Storey,
  Macklin, Hoeller, Rudin, Allshire, Handa, et~al.]{isaacgym}
Viktor Makoviychuk, Lukasz Wawrzyniak, Yunrong Guo, Michelle Lu, Kier Storey,
  Miles Macklin, David Hoeller, Nikita Rudin, Arthur Allshire, Ankur Handa,
  et~al.
\newblock Isaac gym: High performance gpu-based physics simulation for robot
  learning.
\newblock \emph{arXiv preprint arXiv:2108.10470}, 2021.

\bibitem[Moisio et~al.(2013)Moisio, Le{\'o}n, Korkealaakso, and
  Morales]{moisio2013model}
Sami Moisio, Beatriz Le{\'o}n, Pasi Korkealaakso, and Antonio Morales.
\newblock Model of tactile sensors using soft contacts and its application in
  robot grasping simulation.
\newblock \emph{Robotics and Autonomous Systems}, 61\penalty0 (1):\penalty0
  1--12, 2013.

\bibitem[Molchanov et~al.(2016)Molchanov, Kroemer, Su, and
  Sukhatme]{molchanov2016contact}
Artem Molchanov, Oliver Kroemer, Zhe Su, and Gaurav~S Sukhatme.
\newblock Contact localization on grasped objects using tactile sensing.
\newblock In \emph{IEEE/RSJ International Conference on Intelligent Robots and
  Systems (IROS)}, 2016.

\bibitem[Morgan et~al.(2022)Morgan, Hang, Wen, Bekris, and
  Dollar]{dexhand_4_morgan2022complex}
Andrew~S. Morgan, Kaiyu Hang, Bowen Wen, Kostas Bekris, and Aaron~M. Dollar.
\newblock Complex in-hand manipulation via compliance-enabled finger gaiting
  and multi-modal planning.
\newblock \emph{IEEE Robotics and Automation Letters}, 7\penalty0 (2):\penalty0
  4821--4828, 2022.
\newblock \doi{10.1109/LRA.2022.3145961}.

\bibitem[Murali et~al.(2020)Murali, Li, Gandhi, and Gupta]{murali2020learning}
Adithyavairavan Murali, Yin Li, Dhiraj Gandhi, and Abhinav Gupta.
\newblock Learning to grasp without seeing.
\newblock In \emph{Proceedings of the 2018 International Symposium on
  Experimental Robotics}, pages 375--386. Springer, 2020.

\bibitem[Nair and Hinton(2010)]{relu}
Vinod Nair and Geoffrey~E Hinton.
\newblock Rectified linear units improve restricted boltzmann machines.
\newblock In \emph{International Conference on Machine Learning (ICML)}, 2010.

\bibitem[Navarro et~al.(2015)Navarro, Kumar, Fonte, Fraisse, Poisson, and
  Cherubini]{navarro2015active}
Benjamin Navarro, Prajval Kumar, Aicha Fonte, Philippe Fraisse, G{\'e}rard
  Poisson, and Andrea Cherubini.
\newblock Active calibration of tactile sensors mounted on a robotic hand.
\newblock In \emph{Intelligent RObots and Systems Workshop on Multimodal
  sensor-based robot control for HRI and soft manipulation}, 2015.

\bibitem[Okamura et~al.(2000)Okamura, Smaby, and Cutkosky]{dex_overview}
Allison~M Okamura, Niels Smaby, and Mark~R Cutkosky.
\newblock An overview of dexterous manipulation.
\newblock In \emph{IEEE International Conference on Robotics and Automation.
  Symposia Proceedings}, 2000.

\bibitem[Padmanabha et~al.(2020)Padmanabha, Ebert, Tian, Calandra, Finn, and
  Levine]{padmanabha2020omnitact}
Akhil Padmanabha, Frederik Ebert, Stephen Tian, Roberto Calandra, Chelsea Finn,
  and Sergey Levine.
\newblock Omnitact: A multi-directional high-resolution touch sensor.
\newblock In \emph{2020 IEEE International Conference on Robotics and
  Automation (ICRA)}, pages 618--624. IEEE, 2020.

\bibitem[Pan et~al.(2022)Pan, Lepert, Yuan, Antonova, and
  Bohg]{roller_tactile2022}
Chaoyi Pan, Marion Lepert, Shenli Yuan, Rika Antonova, and Jeannette Bohg.
\newblock Task-driven in-hand manipulation of unknown objects with tactile
  sensing.
\newblock \emph{arXiv preprint arXiv:2210.13403}, 2022.

\bibitem[Patel et~al.(2022)Patel, Wang, Radosavovic, and
  Malik]{patel2022learning}
Austin Patel, Andrew Wang, Ilija Radosavovic, and Jitendra Malik.
\newblock Learning to imitate object interactions from internet videos.
\newblock \emph{arXiv preprint arXiv:2211.13225}, 2022.

\bibitem[Petrovskaya and Khatib(2011)]{bin_contact_1}
Anna Petrovskaya and Oussama Khatib.
\newblock Global localization of objects via touch.
\newblock \emph{IEEE Transactions on Robotics}, 27\penalty0 (3):\penalty0
  569--585, 2011.

\bibitem[Pitz et~al.(2023)Pitz, R{\"o}stel, Sievers, and B{\"a}uml]{Pitz2023}
Johannes Pitz, Lennart R{\"o}stel, Leon Sievers, and Berthold B{\"a}uml.
\newblock Dextrous tactile in-hand manipulation using a modular reinforcement
  learning architecture.
\newblock In \emph{IEEE International Conference on Robotics and
  Automation~(ICRA)}, 2023.

\bibitem[Qi et~al.(2022)Qi, Kumar, Calandra, Ma, and Malik]{qi2022hand}
Haozhi Qi, Ashish Kumar, Roberto Calandra, Yi~Ma, and Jitendra Malik.
\newblock In-hand object rotation via rapid motor adaptation.
\newblock In \emph{Conference on Robot Learning~(CoRL)}, 2022.

\bibitem[Qin et~al.(2022{\natexlab{a}})Qin, Huang, Yin, Su, and
  Wang]{qin2022dexpoint}
Yuzhe Qin, Binghao Huang, Zhao-Heng Yin, Hao Su, and Xiaolong Wang.
\newblock Dexpoint: Generalizable point cloud reinforcement learning for
  sim-to-real dexterous manipulation.
\newblock In \emph{Conference on Robot Learning~(CoRL)}, 2022{\natexlab{a}}.

\bibitem[Qin et~al.(2022{\natexlab{b}})Qin, Su, and Wang]{qin2022one}
Yuzhe Qin, Hao Su, and Xiaolong Wang.
\newblock From one hand to multiple hands: Imitation learning for dexterous
  manipulation from single-camera teleoperation.
\newblock \emph{IEEE Robotics and Automation Letters}, 7\penalty0 (4):\penalty0
  10873--10881, 2022{\natexlab{b}}.

\bibitem[Qin et~al.(2022{\natexlab{c}})Qin, Wu, Liu, Jiang, Yang, Fu, and
  Wang]{qin2022dexmv}
Yuzhe Qin, Yueh-Hua Wu, Shaowei Liu, Hanwen Jiang, Ruihan Yang, Yang Fu, and
  Xiaolong Wang.
\newblock Dexmv: Imitation learning for dexterous manipulation from human
  videos.
\newblock In \emph{European Conference on Computer Vision~(ECCV)},
  2022{\natexlab{c}}.

\bibitem[Rajeswaran et~al.(2018)Rajeswaran, Kumar, Gupta, Vezzani, Schulman,
  Todorov, and Levine]{dapg}
Aravind Rajeswaran, Vikash Kumar, Abhishek Gupta, Giulia Vezzani, John
  Schulman, Emanuel Todorov, and Sergey Levine.
\newblock Learning complex dexterous manipulation with deep reinforcement
  learning and demonstrations.
\newblock In \emph{Robotics: Science and Systems~(RSS)}, 2018.

\bibitem[Rus(1999)]{rus1999hand}
Daniela Rus.
\newblock In-hand dexterous manipulation of piecewise-smooth 3-d objects.
\newblock \emph{The International Journal of Robotics Research~(IJRR)},
  18\penalty0 (4):\penalty0 355--381, 1999.

\bibitem[Schulman et~al.(2015)Schulman, Moritz, Levine, Jordan, and
  Abbeel]{schulman2015high}
John Schulman, Philipp Moritz, Sergey Levine, Michael Jordan, and Pieter
  Abbeel.
\newblock High-dimensional continuous control using generalized advantage
  estimation.
\newblock \emph{arXiv preprint arXiv:1506.02438}, 2015.

\bibitem[Schulman et~al.(2017)Schulman, Wolski, Dhariwal, Radford, and
  Klimov]{ppo}
John Schulman, Filip Wolski, Prafulla Dhariwal, Alec Radford, and Oleg Klimov.
\newblock Proximal policy optimization algorithms.
\newblock \emph{arXiv preprint arXiv:1707.06347}, 2017.

\bibitem[Shi et~al.(2017)Shi, Woodruff, Umbanhowar, and Lynch]{dex_6}
Jian Shi, J.~Zachary Woodruff, Paul~B. Umbanhowar, and Kevin~M. Lynch.
\newblock Dynamic in-hand sliding manipulation.
\newblock \emph{IEEE Transactions on Robotics}, 33\penalty0 (4):\penalty0
  778--795, 2017.

\bibitem[Si and Yuan(2022)]{si2022taxim}
Zilin Si and Wenzhen Yuan.
\newblock Taxim: An example-based simulation model for gelsight tactile
  sensors.
\newblock \emph{IEEE Robotics and Automation Letters}, 7\penalty0 (2):\penalty0
  2361--2368, 2022.

\bibitem[Sievers et~al.(2022)Sievers, Pitz, and B{\"a}uml]{Sievers2022}
Leon Sievers, Johannes Pitz, and Berthold B{\"a}uml.
\newblock Learning purely tactile in-hand manipulation with a torque-controlled
  hand.
\newblock In \emph{IEEE International Conference on Robotics and
  Automation~(ICRA)}, 2022.

\bibitem[Smith et~al.(2021)Smith, Meger, Pineda, Calandra, Malik,
  Romero~Soriano, and Drozdzal]{smith2021active}
Edward Smith, David Meger, Luis Pineda, Roberto Calandra, Jitendra Malik,
  Adriana Romero~Soriano, and Michal Drozdzal.
\newblock Active 3d shape reconstruction from vision and touch.
\newblock \emph{Advances in Neural Information Processing Systems},
  34:\penalty0 16064--16078, 2021.

\bibitem[Sodhi et~al.(2022)Sodhi, Kaess, Mukadanr, and
  Anderson]{object_track2022patchgraph}
Paloma Sodhi, Michael Kaess, Mustafa Mukadanr, and Stuart Anderson.
\newblock Patchgraph: In-hand tactile tracking with learned surface normals.
\newblock In \emph{International Conference on Robotics and Automation (ICRA)},
  2022.

\bibitem[Suresh et~al.(2021)Suresh, Bauza, Yu, Mangelson, Rodriguez, and
  Kaess]{tactile_slam}
Sudharshan Suresh, Maria Bauza, Kuan-Ting Yu, Joshua~G Mangelson, Alberto
  Rodriguez, and Michael Kaess.
\newblock Tactile slam: Real-time inference of shape and pose from planar
  pushing.
\newblock In \emph{IEEE International Conference on Robotics and Automation
  (ICRA)}, 2021.

\bibitem[Suresh et~al.(2022)Suresh, Si, Anderson, Kaess, and
  Mukadam]{suresh2022midastouch}
Sudharshan Suresh, Zilin Si, Stuart Anderson, Michael Kaess, and Mustafa
  Mukadam.
\newblock Midastouch: Monte-carlo inference over distributions across sliding
  touch.
\newblock In \emph{Conference on Robot Learning~(CoRL)}, 2022.

\bibitem[Taylor et~al.(2022)Taylor, Dong, and Rodriguez]{taylor2022gelslim}
Ian~H Taylor, Siyuan Dong, and Alberto Rodriguez.
\newblock Gelslim 3.0: High-resolution measurement of shape, force and slip in
  a compact tactile-sensing finger.
\newblock In \emph{International Conference on Robotics and Automation (ICRA)},
  2022.

\bibitem[Tian et~al.(2019)Tian, Ebert, Jayaraman, Mudigonda, Finn, Calandra,
  and Levine]{manipulation_by_feel}
Stephen Tian, Frederik Ebert, Dinesh Jayaraman, Mayur Mudigonda, Chelsea Finn,
  Roberto Calandra, and Sergey Levine.
\newblock Manipulation by feel: Touch-based control with deep predictive
  models.
\newblock In \emph{International Conference on Robotics and Automation (ICRA)},
  2019.

\bibitem[Tobin et~al.(2017)Tobin, Fong, Ray, Schneider, Zaremba, and
  Abbeel]{domain_randomization}
Josh Tobin, Rachel Fong, Alex Ray, Jonas Schneider, Wojciech Zaremba, and
  Pieter Abbeel.
\newblock Domain randomization for transferring deep neural networks from
  simulation to the real world.
\newblock In \emph{IEEE/RSJ international conference on intelligent robots and
  systems (IROS)}, 2017.

\bibitem[Tournassoud et~al.(1987)Tournassoud, Lozano-P{\'e}rez, and
  Mazer]{dex_8}
Pierre Tournassoud, Tom{\'a}s Lozano-P{\'e}rez, and Emmanuel Mazer.
\newblock Regrasping.
\newblock In \emph{IEEE International Conference on Robotics and
  Automation~(ICRA)}, 1987.

\bibitem[Van~Hoof et~al.(2016)Van~Hoof, Chen, Karl, van~der Smagt, and
  Peters]{visiotactile}
Herke Van~Hoof, Nutan Chen, Maximilian Karl, Patrick van~der Smagt, and Jan
  Peters.
\newblock Stable reinforcement learning with autoencoders for tactile and
  visual data.
\newblock In \emph{IEEE/RSJ international conference on intelligent robots and
  systems (IROS)}, 2016.

\bibitem[Wu et~al.(2022)Wu, Wang, and Wang]{wu2023learning}
Yueh-Hua Wu, Jiashun Wang, and Xiaolong Wang.
\newblock Learning generalizable dexterous manipulation from human grasp
  affordance.
\newblock In \emph{Conference on Robot Learning~(CoRL)}, 2022.

\bibitem[Xu et~al.()Xu, Kim, Chen, Garcia, Agrawal, Matusik, and
  Sueda]{tactile_sim}
Jie Xu, Sangwoon Kim, Tao Chen, Alberto~Rodriguez Garcia, Pulkit Agrawal,
  Wojciech Matusik, and Shinjiro Sueda.
\newblock Efficient tactile simulation with differentiability for robotic
  manipulation.
\newblock In \emph{Conference on Robot Learning~(CoRL)}.

\bibitem[Xu et~al.(2022)Xu, Song, and Ciocarlie]{tandem}
Jingxi Xu, Shuran Song, and Matei Ciocarlie.
\newblock Tandem: Learning joint exploration and decision making with tactile
  sensors.
\newblock \emph{IEEE Robotics and Automation Letters}, 7\penalty0 (4):\penalty0
  10391--10398, 2022.

\bibitem[Ye et~al.(2022)Ye, Wang, Huang, Qin, and Wang]{ye2022learning}
Jianglong Ye, Jiashun Wang, Binghao Huang, Yuzhe Qin, and Xiaolong Wang.
\newblock Learning continuous grasping function with a dexterous hand from
  human demonstrations.
\newblock \emph{arXiv preprint arXiv:2207.05053}, 2022.

\bibitem[Yuan et~al.(2017)Yuan, Dong, and Adelson]{yuan2017gelsight}
Wenzhen Yuan, Siyuan Dong, and Edward~H Adelson.
\newblock Gelsight: High-resolution robot tactile sensors for estimating
  geometry and force.
\newblock \emph{Sensors}, 17\penalty0 (12):\penalty0 2762, 2017.

\bibitem[Zhu et~al.(2022)Zhu, Jain, Tomizuka, and Van~Baar]{zhu2022learning}
Xinghao Zhu, Siddarth Jain, Masayoshi Tomizuka, and Jeroen Van~Baar.
\newblock Learning to synthesize volumetric meshes from vision-based tactile
  imprints.
\newblock In \emph{International Conference on Robotics and Automation (ICRA)},
  2022.

\end{thebibliography}
\newpage
\ \\ \\
\newpage
\appendix

\subsection{System Video Demo}
We provide a video demo of our system at {\color{blue}{\texttt{\url{http://touchdexterity.github.io}}}}. The raw video demo can also be founded in the submitted files.

\subsection{PPO Training Hyperparameters}
We use the proximal policy optimization~(PPO) algorithm to train our control policy. The setup of the PPO algorithm is as follows. We use an advantage clipping coefficient $\epsilon = 0.2$. We use a horizon length of 16, with $\gamma=0.99$ and generalized advantage estimator~(GAE)~\cite{schulman2015high} coefficient $\tau=0.95$. The policy network is a three-layer MLP with ELU activation. Its hidden layer is [512, 256, 256]. The policy network's learning rate is set to 1e-4, with an adaptive KL threshold of 0.02. The value network is a four-layer MLP with ELU activation. Its hidden layer is [512, 512, 256, 256]. The value network's learning rate is set to 5e-4, with an adaptive KL threshold of 0.016.  We normalize the state input, value, and advantage during training. We use a gradient norm of 1.0. The minibatch size is set to 16384.

\subsection{Improving Sim2Real Transfer}
\noindent \textbf{Domain Randomization} We use several domain randomization techniques to improve the Sim2Real transfer. The details are shown in Table~\ref{table:dr}.
\begin{table}[htbp]
\renewcommand\arraystretch{1.05}
\caption{Domain Randomization Setup}
\centering
\begin{tabular*}{0.87\linewidth}{l@{\extracolsep{\fill}}c}
\toprule
Object: Mass~(kg)                & [0.2, 0.6]    \\
Object: Friction                   & [0.3, 3.0]     \\
Object: Shape              & $\times\mathcal{U}(0.95, 1.05)$     \\
Object: Initial Position~(cm) &  $+\mathcal{U}(-0.015, 0.015)$ \\
Hand: Friction    & [0.3, 3.0]    \\
\midrule
PD Controller: P Gain         &  $\times\mathcal{U}(0.66, 1.33)$      \\
PD Controller: D Gain     &  $\times\mathcal{U}(0.80, 1.20)$     \\
\midrule
Sensor: Lag Probability         & 0.25      \\
Sensor: Drop Rate               & 0.1      \\
\midrule
Random Force: Scale                    & 0.2       \\
Random Force: Probability              & [0.2, 0.25]    \\
Random Force: Decay Coeff. and Interval & 0.99 every 0.1s     \\ 
\midrule
Joint Observation Noise.           & $+\mathcal{U}(-0.05, 0.05)$     \\
Action Noise.    & $+\mathcal{U}(-0.06, 0.06)$   \\
\bottomrule
\end{tabular*}
\label{table:dr}
\end{table}

\noindent \textbf{System Identification} We apply system identification to align the behavior of the PD controller in simulation to that in the real. We tune the PD coefficients to ensure that the responses of the controllers to the impulse and sinusoidal inputs are aligned. We find this step crucial for successful Sim2Real transfer.
\begin{figure}[t]
\centering
\includegraphics[width=0.9\linewidth]{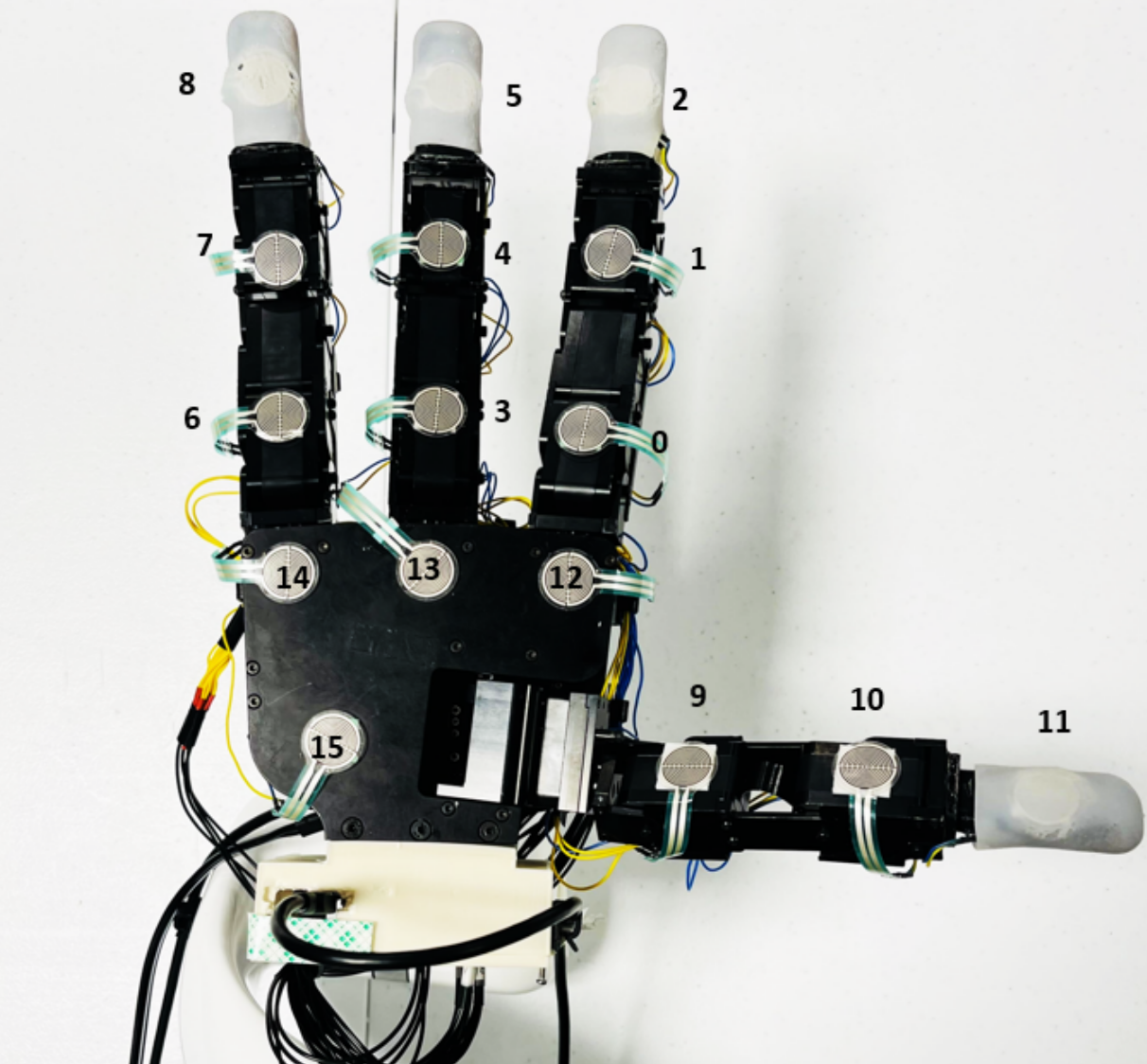}
\caption{Contact sensor map.}
\label{fig:sensor_map}
\vspace{-0.4cm}
\end{figure}
\subsection{Reward Design}
\noindent \textbf{Rotation Reward}
\begin{equation}
    r_{rot} =  {\rm clip} (\Delta \theta, -0.157, 0.157). 
\end{equation}

\noindent \textbf{Velocity Reward}
\begin{equation}
    r_{vel} = -\Vert v_t\Vert.
\end{equation}

\noindent \textbf{Falling Reward~(Penalty)}
\begin{equation}
    r_{fall} = -50.0.
\end{equation}

\noindent \textbf{Work Reward~(Penalty)}
\begin{equation}
    r_{work} = -\langle|\tau|, |\dot{q_t}|\rangle.
\end{equation}

\noindent \textbf{Torque Reward~(Penalty)}
\begin{equation}
    r_{torque} = -\Vert\tau\Vert.
\end{equation}

\noindent \textbf{Distance Reward}
\begin{equation}
    r_{dist} = {\rm mean}_{i=0,1,2,3}({\rm clip} (0.1 /(0.02 + 4d(x_{tip}^i, x_{obj})), 0, 1)).
\end{equation}

\noindent The overall reward function is
\begin{equation}
    r_t = w_1r_{rot} + w_2r_{vel} + w_3r_{fall} + w_4r_{work} + w_5r_{torque} + w_6r_{dist}.
\end{equation}
The setup of each weight: $w_1 = 20.0, w_2 = 0.1, w_3 = 1.0, w_4 = 0.0003, w_5 = 0.0003, w_6=0.1$.

\subsection{More Sensor Response Examples}
\begin{figure*}[t]
\subfigure{\includegraphics[width=\textwidth]{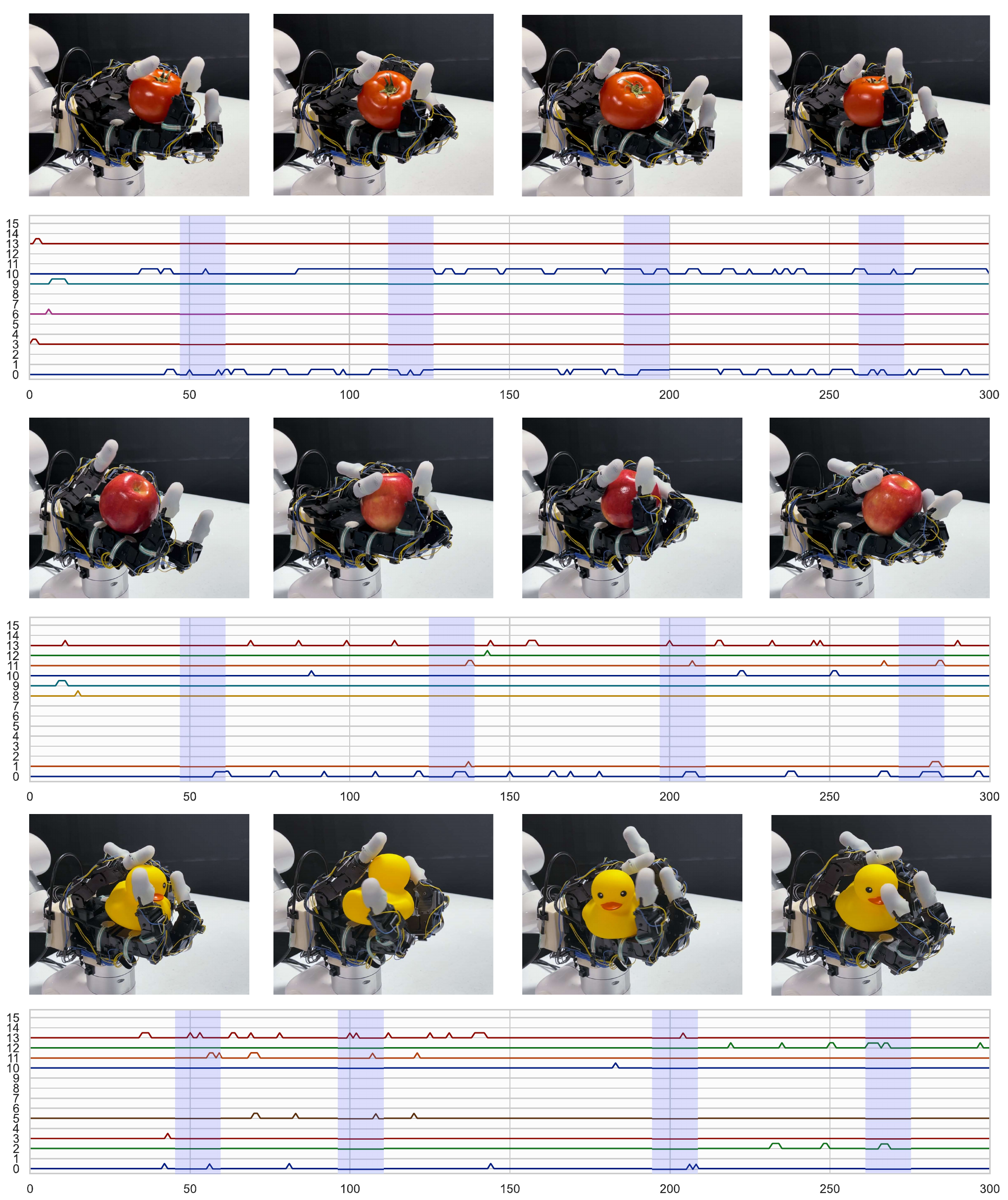}}
\vspace{-0.2in}
\caption{More sensor activation trajectories in the real world experiments. The curves are collected on different objects and display different patterns.}
\label{fig:more_curve}
\vspace{-0.4cm}
\end{figure*}

We show more examples of sensor activation trajectories collected in real-world experiments in Figure~\ref{fig:more_curve}. These trajectories are collected on different objects. We can observe different activation patterns in the trajectories. 

\end{document}